\documentclass[lettersize,journal]{IEEEtran}
\usepackage{amsmath,amsfonts}
\usepackage{subfigure}
\usepackage{algorithmic}
\usepackage{algorithm}
\usepackage{array}
\usepackage[caption=false,font=normalsize,labelfont=sf,textfont=sf]{subfig}
\usepackage{url}
\usepackage{verbatim}
\usepackage{graphicx}
\usepackage{cite}
\graphicspath{{./fig/}}

\usepackage{array}
\usepackage{color}
\usepackage[caption=false,font=normalsize,labelfont=sf,textfont=sf]{subfig}
\usepackage{textcomp}
\usepackage{stfloats}
\usepackage{cleveref}   
\usepackage{multicol}
\usepackage{multirow}
\usepackage{booktabs}
\usepackage{amsmath, amssymb}
\usepackage{makecell}
\begin{document}

\title{Edge-Assisted Multi-Robot Visual-Inertial SLAM with Efficient Communication}

\author{Xin Liu,~\IEEEmembership{Student Member,~IEEE,} Shuhuan Wen,~\IEEEmembership{Senior Member,~IEEE,} Jing Zhao, Tony Z. Qiu,~\IEEEmembership{Member,~IEEE,} Hong Zhang,~\IEEEmembership{Fellow,~IEEE,}
\thanks{\textit{(Corresponding author: Shuhuan Wen)} email: swen@ysu.edu.cn}%
\thanks{The work is supported by the National Natural Science Foundation of China (No. 62273296), China Scholarship Council (No. 202308130063), the Hebei innovation capability improvement plan project(22567619H), the National Natural Science Foundation of China (No. 52172332) and the Key Research and Development Project in Hainan Province (ZDYF2021GXJS015).}
\thanks{Xin Liu, Shuhuan Wen and Jing Zhao are with the Department of Key Lab of Industrial Computer Control Engineering of Hebei Province, Engineering Research Center of the Ministry of Education for Intelligent Control System and Intelligent Equipment, Yanshan University, Qinhuangdao, 066004, China.

Tony Z. Qiu is with Intelligent Transport System Research Center, Wuhan University of Technology,	Wuhan, China, and the Department of Civil and Environmental Engineering at the University of Alberta, Canada.

Hong Zhang is with the Department of Electronic and Electrical Engineering, Southern University of Science and Technology, Shenzhen, China.
}
\thanks{Manuscript received April 19, 2021; revised August 16, 2021.}}

\markboth{Journal of \LaTeX\ Class Files,~Vol.~14, No.~8, August~2021}%
{Shell \MakeLowercase{\textit{et al.}}: A Sample Article Using IEEEtran.cls for IEEE Journals}


\maketitle

\begin{abstract}
The integration of cloud computing and edge computing is an effective way to achieve global consistent and real-time multi-robot Simultaneous Localization and Mapping (SLAM). Cloud computing effectively solves the problem of limited computing, communication and storage capacity of terminal equipment. However, limited bandwidth and extremely long communication links between terminal devices and the cloud result in serious performance degradation of multi-robot SLAM systems. To reduce the computational cost of feature tracking and improve the real-time performance of the robot, a lightweight SLAM method of optical flow tracking based on pyramid IMU prediction is proposed. On this basis, a centralized multi-robot SLAM system based on a robot-edge-cloud layered architecture is proposed to realize real-time collaborative SLAM. It avoids the problems of limited on-board computing resources and low execution efficiency of single robot. In this framework, only the feature points and keyframe descriptors are transmitted and lossless encoding and compression are carried out to realize real-time remote information transmission with limited bandwidth resources. This design reduces the actual bandwidth occupied in the process of data transmission, and does not cause the loss of SLAM accuracy caused by data compression. Through experimental verification on the EuRoC dataset, compared with the current most advanced local feature compression method, our method can achieve lower data volume feature transmission, and compared with the current advanced centralized multi-robot SLAM scheme, it can achieve the same or better positioning accuracy under low computational load.

\textit{Note to Practitioners—}The purpose of this paper is to reduce the communication load of a Cloud-Edge-Robot system by compressing and transmitting of keyframes and non-keyframes, respectively, which is suitable for a multi-robot SLAM system and can realize multi-robot joint localization and sparse map reconstruction under efficient communication. Currently, remote SLAM or centralized multi-robot SLAM is usually implemented by transferring the whole image or the features and descriptors of the image. In this paper, lightweight SLAM optical flow tracking based on pyramid IMU prediction is implemented to track non-keyframes. At the edge server, tracking between non-keyframes is realized only by transmitting keypoints. For keyframes, the pose estimation is realized by transmitting compressed features and descriptors. Multi-robot localization and map fusion are realized in the cloud through key frame feature information. Experiments on public datasets show that this method is feasible and can achieve high-precision joint positioning with a low amount of transmitted data. In future studies, we will apply this framework to more real-world systems, while achieving rich, accurate map fusion with more advanced features.
\end{abstract}


\section{Introduction}
\IEEEPARstart{W}{ith} the improvement of the Internet of Things (IoT) and intelligent systems, collaborative autonomous perception and localization of complex and large-scale scenes are the general trends. SLAM, as a key technology for robot intelligence and autonomy, is developed vigorously in the world in recent years. Researchers are working to apply SLAM technology to real-world applications, such as mobile robots \cite{r1, r2}, vehicles \cite{r3}, and underwater robots \cite{r4}. The IoT combines various sensor devices with the Internet to form a network that extensively connects people, machines and things at any time and place \cite{ref5}. The edge computing can improve the intelligence of the IoT. Edge computing is applied in a wide range of fields in the IoT. It is especially suitable for application scenarios with special service requirements such as low delay, high bandwidth, high reliability, massive connections, heterogeneous aggregation, local security and privacy protection. IoT Edge \cite{ref6} is a set of intelligent systems deployed and running on edge node devices. It extends cloud computing capability to edge nodes and executes intelligent computing on site, which is an important support for distributed autonomy and automatic control.

Typically, SLAM is implemented using two major types of sensors: camera and LiDAR. Compared with SLAM technology using LiDAR, the cameras used in Visual SLAM (especially monocular cameras) are lightweight, low-cost and contain rich visual information, so they are suitable for platforms with limited cost and load. However, since it is difficult for a monocular camera to obtain depth directly from a single image, the scale of the real scene is often not observed. To solve this problem, a monocular visual-inertial odometry can obtain true scale information through alignment vision and inertial odometer measurements. In fact, the monocular visual inertial odometer is the smallest combination kit that can meet the requirements of lightweight and real environment perception \cite{ref7}.

\begin{figure}[h]
	\centering
	\hspace{-4mm}
	\subfigure[Centralized architecture]{
		\includegraphics[width=0.22\textwidth]{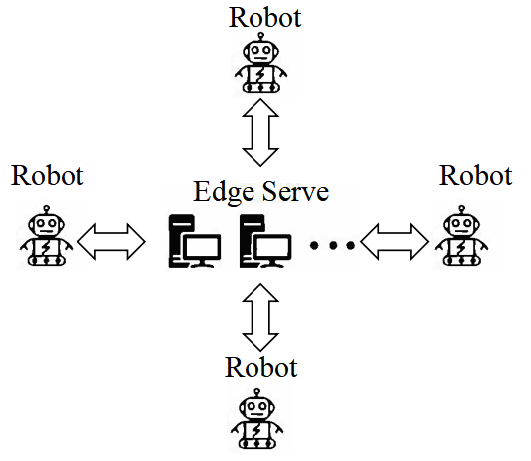}
	}\hspace{-8mm}
	\quad
	\subfigure[Decentralized architecture]{
		\includegraphics[width=0.24\textwidth]{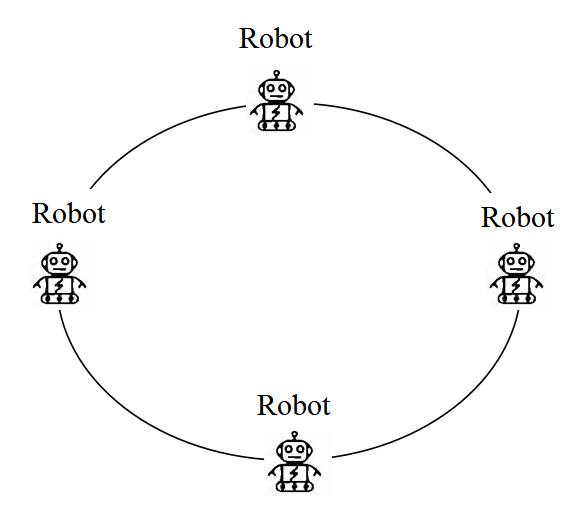}
	}
	\caption{Multi-robot Collaborative SLAM communication architecture.}
	\label{FIG:1}
	\vspace*{-0.5cm}
\end{figure}

Although SLAM has been developing in recent years, it is difficult for a single mobile robot to complete tedious tasks in complex and large scenes. Multi-robot systems improve the performance of a single robot by communicating, collaborating and sharing information, such as task execution efficiency, robustness, flexibility and fault tolerance. Because of the limitations of single-robot SLAM on the task and the strong flexibility and adaptability of multi-robot systems, multi-robot collaborative SLAM has attracted more and more attention. As shown in Fig. \ref{FIG:1}, in multi-robot collaborative SLAM, when building a local map, a single robot needs to integrate all available data to assist it in accurate positioning in the global environment, and carry out data association through communication between multiple robot systems to coordinate and build a consistent global map. Architecture selection and task assignment, relative pose determination, communication and data association, map fusion and back-end optimization are the key problems of multi-robot SLAM. Through the intelligent edge computing of the Internet of Things, the robot can easily access the edge device, and quickly complete the work on the SLAM cloud and the transmission of sensor data.

The distributed multi-robot system shown in Fig. \ref{FIG:1} (b) can usually easily add more robots, while each robot can be calculated in parallel, speeding up the mapping and localization of the entire system, e.g. \cite{ref38,ref39}. However, the distributed multi-robot SLAM system requires a lot of data sharing and communication, which has high requirements on communication efficiency and network load. As shown in Fig. \ref{FIG:1} (a), the centralized multi-robot system can ensure the global consistency of data due to its centralized architecture, which is extremely important for SLAM systems, e.g. \cite{ref46,ref47}. Meanwhile, the central node can balance the computing load through task allocation, which greatly improves the efficiency of the entire system. Whether it is a distributed or centralized system, the efficiency of communication and the amount of data can lead to performance bottlenecks, affecting the real-time and efficiency of the entire system \cite{add1,add2,add3}.

To address all the above issues, we propose a fast, lightweight, real-time multi-robot monocular Visual-Inertial SLAM system. To summarize, our key contributions are as follows:
\begin{itemize}
	\item [1)] 
	We propose a light-weight SLAM method based on the ORB-SLAM3 \cite{ref8} and Lucas Kanade (LK) spare optical flow method \cite{ref9}. In this method, the calculation of descriptors only takes place in keyframes, and the pose estimation between non-keyframes is realized by using the spare optical flow tracking method based on IMU prediction in the pyramid. This design greatly reduces the computational consumption in the process of feature tracking, and does not affect the accuracy of the back-end optimization calculation.    
	\item [2)]
	Based on the above fast SLAM, we propose a centralized multi-robot SLAM system based on the hierarchical architecture of robot-edge-cloud to achieve real-time collaborative SLAM. We decouple the classic SLAM framework and layer feature extraction, Visual-Inertial odometry (VIO), and global optimization. This design avoids the problems of limited onboard computing resources and low execution efficiency of a single robot.
	\item [3)]
	By lossless coding and compression of corner points (non-keyframes) and local binary descriptors (keyframs) in the robot-edge-cloud architecture, we can realize real-time remote information transmission under the condition of limited bandwidth resources. This design reduces the actual bandwidth occupied in the data transmission process and does not cause the loss of SLAM accuracy caused by data compression. In addition, to guarantee the optimization efficiency and reduce the storage, map backbone profiling (MBP) is introduced into our method.
\end{itemize}

\section{Related Work}
In this section, we discuss the related work in the following four branches: 1) Visual-Inertial SLAM (VINS), 2) muti-robot SLAM,and 3) Edge-assisted SLAM.
\subsection{Visual-Inertial SLAM}
For multi-robot SLAM systems, each robot is a single moving individual, the accuracy of a single robot SLAM are equally important. Classical V-SLAM methods are mainly divided into the SLAM based on the direct method \cite{ref10, ref11} and the SLAM based on the feature point method \cite{ref12, ref13, ref14}. Purely V-SLAM methods have developed rapidly in recent years, and here we skip the discussion of them and focus on VINS. According to the optimization method of SLAM, VINS/VIO scoring is divided into two types: one is the filtering method represented by extended Kalman filter (EKF), and the other is the nonlinear optimization method represented by graph optimization.

In the case of limited computing resources, the filtering method is more effective. The Multi-State Constraint Kalman Filter (MSCKF) proposed by Mourikis \textit{et al.} \cite{ref15} is one of the earliest VINS system. MSCKF integrates IMU and visual information under the EKF framework. Compared with the simple VO algorithm, MSCKF has higher robustness. Li \textit{et al.} \cite{ref16}, \cite{ref17} further ensure the correct observability of the established linear system on the basis of EKF, effectively improving the localization accuracy in large scenes. Based on \cite{ref15}, Sun \textit{et al.} \cite{ref18} further extend MSCKF to stereo camera, which improved the robustness of the algorithm and and enable the scale of the scene to be measured. ROVIO \cite{ref19} transforms the photometric error of pixel blocks to obtain the covariance information in EKF, and then updates the filtering state. Based on \cite{ref19}, Michael \textit{et al.} \cite{ref20} further improved the robustness of the algorithm by improving EKF into IEKF. In the latest research, the SLAM method based on EKF is still widely used in the resource-limited scenario, especially in the scheme based on radar sensor for perception \cite{ref21}, \cite{ref22}, \cite{ref23}. However, one of its major disadvantages is that the storage capacity and the amount of states have a square growth relationship, which is not suitable for large-scale scenarios. In V-SLAM, there are many landmarks, so the filtering method is inefficient. However, graph-based optimization SLAM is usually based on keyframe (KF), and the Bundle Adjustment (BA) \cite{ref24} is used to establish the relative transformation relationship between nodes, which ensures accuracy and real-time meanwhile.

Leutenegger \textit{et al}. \cite{ref25} which maintain a sliding window to optimize the state of keyframes in the last period of time, and further optimize frame and IMU jointly. The results show that the proposed has higher accuracy than the loosely-coupled and filter-based SLAM. The VINS-Mono proposed by Qin \textit{et al.} \cite{ref26} is a powerful and universal monocular VI state estimator. Qin \textit{et al.} \cite{ref27} also propose a stereo VINS, which fused stereo camera and IMU with a nonlinear optimization method and significantly improved the accuracy. VI ORB-SLAM \cite{ref28} adds IMU information to assist visual measurement based on \cite{ref13}, \cite{ref14}, and ORB (Oriented FAST and Rotated BRIEF) \cite{ref29} feature is adopted for feature extraction. Based on \cite{ref28}, \cite{ref30}, the ORB-SLAM3 proposed by Campos \textit{et al.} \cite{ref8} is one of the state-of-the-art methods. It can operate stably on monocular, stereo and RGB-D cameras using pinhole or fisheye models, relying only on maximum posterior estimation (including IMU at initialization), and can operate in  indoors or outdoors. Therefore, we adopt ORB-SLAM3 as a single robot SLAM scheme.

\subsection{Muti-Robot SLAM}
The global perception of multi-robot can be realized through map reuse technology, such as \cite{ref8}, \cite{ref26}. However, map reuse requires a prior map, which is not always realized. In addition, multi-robot SLAM can be divided into centralized and decentralized architectures based on the communication architecture.

DDF-SLAM \cite{ref31} is the earliest solution to decentralized multi-robot SLAM. On this basis, Cunningham \textit{et al.} \cite{ref32} fuse local map with neighboring map to avoid the problem of redundant computation in \cite{ref31}. Choudhary \textit{et al.} \cite{ref33}, \cite{ref34} solved the multi-robot collaborative SLAM problem by using decentralized Gauss-Seidel (DGS) algorithm. This method requires less information exchange and has the property of being readily available, which can be well extended to large teams. Based on this, Cieslewski \textit{et al.} \cite{ref35} integrate the DGS algorithm into a complete V-SLAM system. \cite{ref35} achieves relative recognition among robots through the pre-trained NetVLAD model \cite{ref36}. Although NetVLAD can achieve a high accuracy of scene recognition, the model is complicated and it is difficult to meet the real-time requirements. Lajoie \cite{ref37}, Tian \cite{ref38} and Huang \cite{ref39}\textit{et al.} optimize the relative pose of the robot through Pose Graph Optimization (PGO) to achieve higher accuracy than DGS. Distributed multi-robot PGO can be distributed on multiple robots to accelerate the speed of mapping and localization of the whole system, but this may lead to the problem of data asynchronism caused by the asynchronism between robots, thus affecting the accuracy of mapping and localization. At the same time, distributed parallel computing will increase the load of the network. In general, decentralized robots have great scalability, but they have great challenges in ensuring data consistency and avoiding double computation.
\begin{figure*}[htbp]
	\centering
	\includegraphics[width=0.85\textwidth]{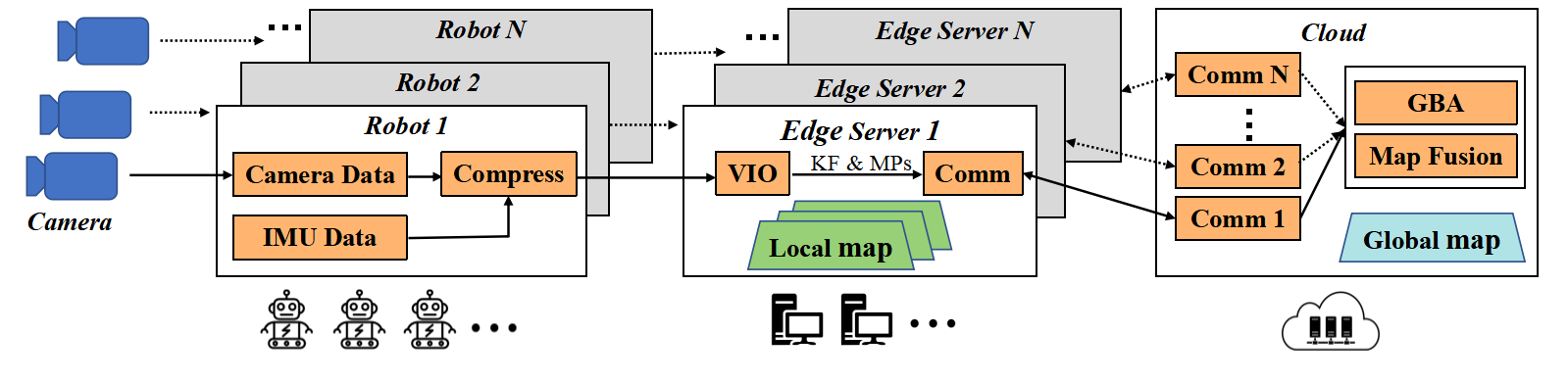}
	\vspace*{-0.25cm}
	\caption{System overview. Each robot is equipped with a Visual-Inertial sensor suite. After analyzing the sensor information, the compressed data is transmitted to a fixed edge server. The edge server can conduct location and map construction in local scenes by running the VIO in real-time and independently. The communication interface is used to exchange data (keyframe local binary features and mappoints) between the edge server and the cloud. The cloud can run computationally expensive, global, and non-real-time tasks: redundant keyframe culling, loop closure, and global map fusion.}
	\label{FIG:2}
	\vspace*{-0.3cm}
	\centering
\end{figure*}
Most collaborative V-SLAM systems use a centralized architecture. CoSLAM \cite{ref40} is a powerful centralized SLAM system that relies only on vision. The system is grouped according to the overlap of frames to process dynamic scenes. Forster \textit{et al.} \cite{ref41} propose a centralized SLAM system based on descriptors, which can complete map fusion by matching descriptors. Bonin-Front \textit{et al.} \cite{ref42} propose a meeting-based centralized system for the difficulty of matching descriptor due to perspective changes. Global information such as GPS can also be used to estimate the relative position of multi-robot, but the introduction of external sensors is greatly limited in the scenario where GPS is unavailable \cite{ref43}. CCM-SLAM \cite{ref44} proposes to effectively utilize the server to assist in solving the high-cost tasks through the two-way communication between the server and the robot. Wen \textit{et al.} \cite{ref45} enhanced the stability of CCM-SLAM under low bandwidth by reducing the amount of data transmitted, but compared with the current advanced SLAM system, it still lacks stable collaborative positioning and more client extensibility. CVI-SLAM \cite{ref46}, COVINS \cite{ref47} and CVIDS \cite{ref48} enhance the collaborative location effect of the system by fusing IMU information, but these systems are built when the communication between the robot and the server is stable. The centralized SLAM system for unstable communication is still to be studied.

\subsection{Edge-Assisted SLAM}
Edge computing \cite{ref49} shifts the workload from the cloud to the server that is physically close to the terminal device. In recent years, researchers have explored the use of edge computation to assist SLAM computation.  Xu \textit{et al.} \cite{ref50} propose an edge-assisted mobile Semantic V-SLAM framework, which can perform V-SLAM in real time. Edge-SLAM \cite{ref51} separates SLAM from a modular perspective. However, these only involve a single robot and an edge node, and cannot be directly extended to the communication between multiple robots and multiple edges. Huang \textit{et al.}\cite{ref52} optimize the edge computing environment by allocating server resources to multiple robots equipped with 2D Lidar, thus accelerating the computing process of SLAM. Lan \textit{et al.} \cite{ref53} accelerated the process of SLAM by unloading different threads to heterogeneous edge servers to achieve the effect of real-time operation, but this has many restrictions on the structure of SLAM system and the architecture of threads. Although these systems accelerate SLAM via edge computing, most of them focus on single robot scenarios. At the same time, there is still room for research on the communication bandwidth and communication stability of edge SLAM.

There are two different approaches to sensor information in edge SLAM: Compression and then Analysis (CTA), where visual information is transmitted as a compressed video sequence. Analysis followed by compression (ATC), in which visual cues are extracted, compressed, and transmitted in the form of local features. For example, Wang \textit{et al.} \cite{ref54} build a remote SLAM system to transmit complete image. However, the transmission of the complete image occupies a higher bandwidth. Van \textit{et al.} \cite{ref55}, \cite{ref56} reduce the bandwidth of information by compressing transmission after analyzing the image. But typically evaluating keypoint descriptors in indirect methods is time consuming, and these descriptors are not reused except by using keyframes\cite{ref57}. The calculation of non-keyframe descriptors can cause time delays in the Edge SLAM system.

\section{System Overview}
The system architecture of the proposed is shown in Fig. \ref{FIG:2}. We propose a new architecture for a centralized VIO. This is done primarily through three layers: the robot's end, the edge, and the cloud. This architecture is first introduced in \cite{ref51} to run multi-robot SLAM equipped with 2D LiDAR sensors. Since real-time, high-resolution image data tend to occupy a large memory, there are difficulties in practical multi-robot applications, transmission and analysis. In this work, we extend this architecture to modes where Visual-Inertial is available by analyzing then compressing sensor data. On the robot, we adopt fast feature detection and feature matching based on the IMU-assisted LK method for non-keyframes (see Section 4), and extract ORB features for keyframes. In the Edge, the VIO front-end is run and the pose of the current frame is solved through the matching result of optical flow tracking in the non-keyframe. For the keyframe, when the optical flow tracking fails, the tracking method in ORB-SLAM3 is adopted for positioning. Meanwhile, the edge maintains a local map of limited size for each agent to ensure the basic autonomy of each agent. At the same time, global map maintenance and computation-heavy processes (PGO and GBA) are moved to Cloud (see Section 6). We use a common communication interface between the robot, the edge and the cloud to realize collaboration between multiple agents. To compress the amount of data transmitted, we encode keyframes and non-keyframes separately through post-analysis compression (see Section 5).
\section{IMU-assisted non-keyframes feature matching}
For feature matching between two consecutive frames, one way is to calculate descriptors to match keypoints between adjacent images, such as \cite{ref12}. However, the computation cost of descriptor calculation and matching is large. In practice, Scale-Invariant Feature Transform (SIFT) cannot be extracted in real-time on the CPU, and the extraction of ORB features takes nearly 20ms, which is a challenge to the computing power of the onboard computer. The other is to minimize the photometric error by the direct method. The method of feature matching by using minimizing the photometric error can only be applied to similar frame to frame matching \cite{ref58}. Therefore, this method usually does not include loop detection. To complete the loop detection and map fusion among multi-robot while improving the computational efficiency, in our method, the LK sparse optical flow tracking is carried out between non-keyframes to complete the inter-frame tracking, while the descriptor calculation only occurs in the keyframe.

The sparse optical flow method can quickly estimate the correspondence between adjacent frames, but it is based on the following three assumptions: 1) the luminosity between adjacent frames is constant; 2) The movement of objects between adjacent frames is small; 3) Adjacent pixels have the same motion. The constant luminosity hypothesis for the $i\rm{th}$ and the $j\rm{th}$ frame can be expressed as:
\begin{align}
{\rm{I(}}{u_1}, {v_1}, i{\rm{)  =  I(}}{u_2}, {v_2}, j{\rm{) }},
\end{align}
where $(u_1, v_1)$ and $(u_2, v_2)$ are the matching points in the $i\rm{th}$ and the $j\rm{th}$ frame, and $\rm I$ is the gray value of the corresponding point.

\begin{figure}[h]
	\centering
	\includegraphics[width=0.4\textwidth]{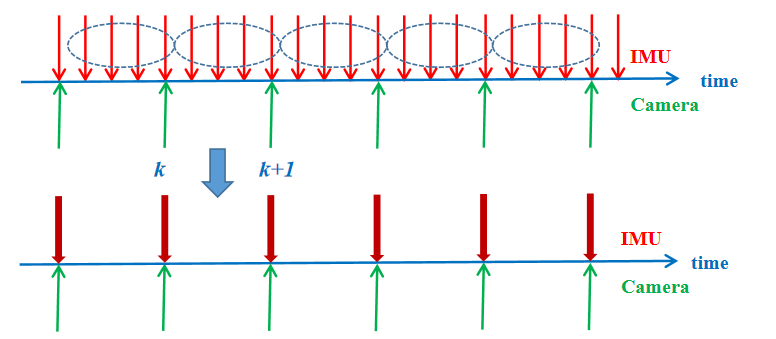}

	\vspace*{-0.25cm}
	\caption{IMU pre-integration diagram. Integrating all IMU information between the $ k\rm {th} $ frame and the $ (k+1) \rm {th} $ frame, the position, velocity and rotation (PVQ) of the  $ (k+1) \rm {th} $ frame can be obtained as the initial value of the visual estimation.}
	\label{FIG:3}
	\centering
\end{figure}

In the real scene, the gray value of the corresponding point will change with the difference of Angle and exposure degree. To reduce the influence caused by the onstant luminosity hypothesis, FastORB-SLAM \cite{ref57} improves the matching accuracy by establishing a uniform acceleration motion (UAM) model. UAM model can predict the position of key points through the information of external sensor like camera. In static scene, compared with uniform motion model (UMM), \cite{ref14} can improve the matching accuracy. However, in dynamic scenes, external sensors are prone to cause unreliable estimates, and the accuracy of UAW models calculated by camera will also be affected. As shown in Fig. \ref{FIG:3}, we introduces external sensors (IMU) into our system, which can provide accurate and fast pose prediction through pre-integration of IMU. Meanwhile, internal sensors are not affected by dynamic objects, which can improve the robustness of the system.

Before the initialization of IMU was completed, we used the uniform motion model to estimate the pose of the current frame through the motion of the past consecutive IMU information. Unlike \cite{ref14}, we calculate the velocity in the UMM model by calculating the mean angular velocity in the IMU frame. For the $K$ sets of velocity data between the $i\rm{th}$ and the $j\rm{th}$ frame, the mean angular velocity can be calculated by:

\begin{align}
{{\bf{v}}_{mean}} = \frac{{\sum\limits_{k = 1}^K {{{\bf{v}}_i}} }}{{{t_i} - {t_j}}},s.t{t_i} < {t_k} < {t_j},
\end{align}
where ${t_i}$ and ${t_j}$ are the timestamps of frame $i$ and frame $j$, respectively, and ${t_k}$ is the timestamp of IMU. Here, since the translation vector is negligible for a short period of time, we only consider the rotational angular velocity. When the IMU is initialized, the pose of the current frame is estimated by IMU pre-integration. The initialization of IMU is obtained by the Maximum-a-Posterior estimation (MPA) \cite{ref8}. 

Considering the IMU consisting of a triaxial accelerometer and a triaxial gyroscope, the measured values of acceleration ${{\hat \omega }_B}$ and angular velocity ${{\hat {\rm a}}_B}$ are affected by white noise $\eta$ and sensor bias $\rm b$. Its model can be described as:
\begin{align}
\begin{array}{l}
{{\hat \omega }_B}(t) = {\omega _B}(t) + {{\mathop{\rm b}\nolimits} ^g}(t) + {\eta ^g}(t)\\
{{\hat {\rm a}}_B}(t) = {\rm{R}}_B^W({{\mathop{\rm a}\nolimits} _B}(t) - {g_W}) + {{\mathop{\rm b}\nolimits} ^a}(t) + {\eta ^a}(t)
\end{array},
\end{align}
where ${( \cdot  )_B}$ and ${( \cdot  )_W}$ represent the body frame and the world frame, respectively. ${\omega _B}$ and ${\mathop{\rm a}\nolimits} _B$ represent the real angular velocity and acceleration in the world frame. ${g_W}$ is the gravitational acceleration in the world frame. $\eta ^g$ and $\eta ^a$ are gyroscope noise and accelerometer noise with mean 0, respectively. ${\mathop{\rm b}\nolimits} ^g$ and ${\mathop{\rm b}\nolimits} ^a$ are gyroscope bias and accelerometer bias, respectively. And their random walk errors are usually modeled using the Wiener process. Formulate the IMU preintegration on a manifold \cite{ref59}, The rotation increment $\Delta {{\rm{\tilde R}}_{ij}}$, velocity increment $\Delta {{\rm{ \tilde v}}_{ij}}$ and position increment $\Delta {{\rm{\tilde p}}_{ij}}$ between the $i\rm{th}$ and the $j\rm{th}$ frame can be expressed as:
\begin{align}
\begin{array}{l}
\Delta {{\rm{\tilde R}}_{ij}} = \prod\limits_{k = i}^{j - 1} {{\mathop{\rm Exp}\nolimits} (\hat \omega ({t_k}) - {{\rm{b}}^g}({t_k}) - {\eta ^g}({t_k})\Delta {t_{ij}})} \\
\Delta {{\mathop{\rm {\tilde v}}\nolimits} _{ij}} = \sum\limits_{k = i}^{j - 1} {\Delta {{\rm{R}}_{ik}}(\hat a({t_k}) - {{\rm{b}}^a}({t_k}) - {\eta ^a}({t_k}))} \Delta {t_{ij}}\\
\Delta {{\mathop{\rm {\tilde p}}\nolimits} _{ij}} = \sum\limits_{k = i}^{j - 1} {\left[ {\Delta {{\mathop{\rm v}\nolimits} _{ik}}\Delta {t_{ij}} + \Delta {{\rm{R}}_{ik}}(\hat a({t_k}) - {{\rm{b}}^a}({t_k}) - {\eta ^a}({t_k}))\Delta t_{ij}^2} \right]} 
\end{array}.
\end{align}
With the benefit of IMU assisted pose estimation, we can accurately predict the position of keypoints and complete sparse optical flow tracking between non-key frames without calculating descriptors. The complete steps are summarized in Algorithm \ref{alg1}.
\begin{algorithm} 
	\caption{IMU-assisted LK tracking} 
	\label{alg1} 
	\begin{algorithmic}
		\REQUIRE Reference Frame: $\rm I_r$, current frame: $\rm I_c$, keypoints in $\rm I_r$: $kps_r$, observed value of IMU: $gyr$, $acc$;
		\STATE The pyramid with scale ratio $s=1.2$, and number of layers $n=8$ is established to describe $\rm I_r$ and $\rm I_r$;
		\IF{IMU is not initialized} 
		\STATE {${{\bf{v}}_{mean}} \longleftarrow $ The average velocity between $\rm I_r$ and $\rm I_c$ is estimated by equation (2);}
		\STATE ${{\rm{T}}_{rc}} \longleftarrow $ The transformation matrix from $\rm I_r$ to $\rm I_c$ is estimated by the UMM;
		\RETURN ${\rm{T}}_{rc}$;
		\ELSIF{IMU is initialized} 
		\STATE{$\Delta{\rm{R}},\Delta{\rm{v}},\Delta{\rm{p}}\longleftarrow $  Calculates the IMU pre-integration by equation (4);} \\
    	\STATE{${{\rm{T}}_{rc}} \longleftarrow $ The transformation matrix from $\rm I_r$ to $\rm I_c$ is estimated by the $\Delta{\rm{\tilde R}},\Delta{\rm{\tilde v}},\Delta{\rm{\tilde p}}$;}
    	\RETURN ${\rm{T}}_{rc}$;
		\ENDIF
		\STATE $kps_{pred} \longleftarrow$ ${\rm{T}}_{rc}$ is used to predict the pixel coordinates corresponding to $kps_r$ in $\rm I_c$;\\
		\STATE $pts_{L8}\longleftarrow$ On the basis of $kps_{pred}$, the LK tracking of the top pyramid is carried out between $\rm I_r$ and $\rm I_c$;
		\STATE $pts_{L1}\longleftarrow$ Iterate $pts_{L8}$ down the pyramid until you reach level 1;
		\STATE Screen the corresponding points by the rotation integral of IMU;
		\STATE $ kps_c$,  $kps_r\longleftarrow$ Eliminate outliners by RANSAC; 
		\STATE Solve pose matrix of $\rm I_c$ in the world frame by PNP;
		\ENSURE The final matched points $ kps_c$,  $kps_r$, the pose matrix of $\rm I_c$ in the world frame: $ T_{cw} $;
	\end{algorithmic} 
\end{algorithm}
\section{Method of communication}
Benefit from the IMU-assisted non-keyframes feature matching method (Section \uppercase\expandafter{\romannumeral3}), in our proposed cloud-edge layered architecture, we can achieve accurate positioning and sparse map establishment without transferring non-keyframe descriptors. In this mode, we can accomplish multi-robot SLAM with minimal information transfer.
\subsection{Feature Coding Compression}
\begin{figure}[h]
	\centering
	\includegraphics[width=0.45\textwidth]{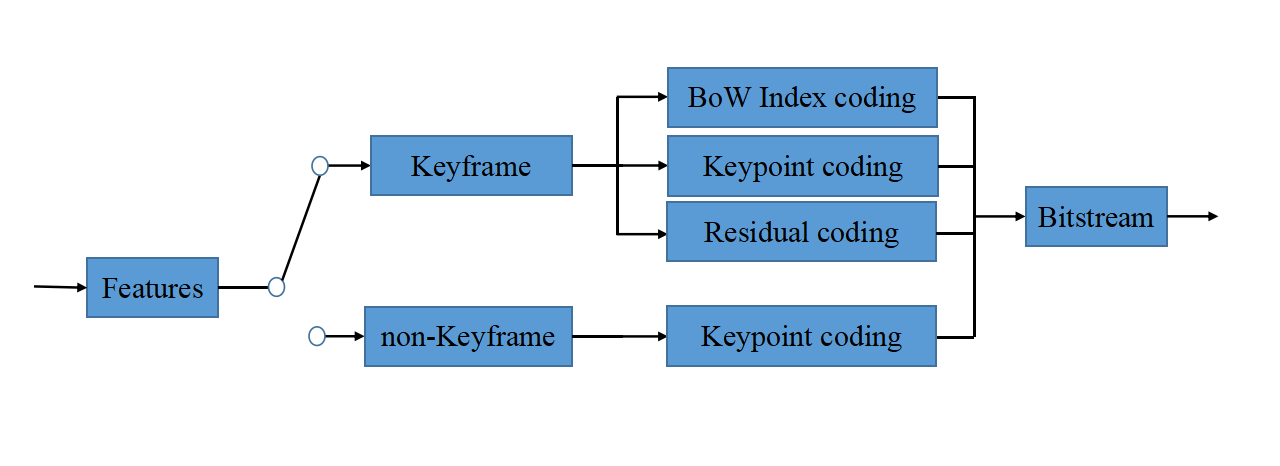}
	\caption{The choice of encoding mode. The decision of keyframes depends on the number of points tracked by the IMU-assisted LK optical flow and the time interval from the previous keyframe.}
	\label{FIG:4}
	\vspace*{-0.25cm}
	\centering
\end{figure}

Based on the feature-compression-based remote SLAM system proposed in \cite{ref55}, we improve the feature-compression method and build a feature-compression-based multi-robot SLAM system. As shown in Fig. \ref{FIG:4}, in order to achieve lossless compression, we adopt the mode of intra-frame encoding, and further divide the mode of intra-frame encoding into keyframe mode and non-keyframe mode. In the non-keyframe mode, we encode and transmit only the keypoints. In the keyframe mode, we compress and encode the visual word corresponding to ORB features, keypoints of ORB features, and descriptors of ORB features. Therefore, the cost of encoding feature $n$ in the $i\rm {th}$ non-keyframes can be expressed as:
\begin{align}
C_{i,n}^{non -kf} = C_{i,n}^{non -kf,kp},
\end{align}
and, the cost of encoding feature $n$ in the $i\rm {th}$ keyframes can be expressed as:
\begin{align}
C_{i,n}^{kf} = C_{i,n}^{kf,kp} + C_{i,n}^{kf,BoW} + C_{i,n}^{kf,des},
\end{align}
where, $C_{i,n}^{non-kf,kp}$, $C_{i,n}^{kf,kp}$ represents the encoding cost of non-keyframe and keyframe key points, respectively, $C_{i,n}^{kf,BoW}$ represents the encoding cost of the visual word index, $C_{i,n}^{kf,des}$ represents the encoding cost of descriptors.
\subsubsection{Compression encoding of keypoint}
In the non-keyframe, we define the keypoint information that only needs to have the horizontal and vertical coordinates and the hierarchy of the pyramid in which they are located. Therefore, a keypoint $n$ in the non-keyframe $i$ can be described as:
\begin{align}
kp_{i,n}^{non-kf} = [u,v,\sigma],
\end{align}
where, $u$ and $v$ represent the horizontal and vertical coordinates in the image coordinate frame, $\sigma$ represents the pyramid level where the key point is located. In the keyframe, in order to make the feature points robust to rotation, we define the keypoints as an array with 4 properties as:
\begin{align}
kp_{i,n}^{kf} = [u,v,\sigma,\theta],
\end{align}
where $\theta$ represents the direction of keypoints calculated by the gray centroid method. $N_\theta$ means divide $2\pi$ radians into $N_\theta$ parts and number each part counterclockwise from smallest to largest. Therefore, the code of keypoints can be converted into the code of $u$, $v$, $\sigma$, $\theta$. In order to further reduce the bandwidth occupancy, we use the horizontal coordinate $u_\sigma$ and vertical coordinate $v_\sigma$ in the pyramid coordinate frame to replace $u$ and $v$.Therefore, the number of bits required to compress keypoint $n$ on non-keyframe $i$ is:
\begin{align}
C_{i,n}^{non-kf,kp} = {\log _2}({N_\sigma }) + {\log _2}(H(\sigma )) + {\log _2}(W(\sigma )),
\end{align}
and the number of bits required to compress keypoint $n$ on keyframe $i$ is:
\begin{align}
C_{i,n}^{kf,kp} = {\log _2}({N_\sigma }) + {\log _2}({N_\theta }) + {\log _2}(H(\sigma )) + {\log _2}(W(\sigma ))
\end{align}
where ${N_\sigma }$ is the number of layers in the pyramid and is assigned the value 8, and ${N_\theta }=32$. $H( \cdot )$ and $W( \cdot )$ are the height and width of the corresponding pyramid, respectively.
\subsubsection{Compression encoding of visual word index and descriptor}
Use the calculated visual vocabulary in ORB-SLAM3 as a reference to the feature coding, and use the visual vocabulary list as shared information that both the robot and the edge can access. For a feature $n$ in the image $i$, there exists a word $v$ with the smallest hamming distance from $n$ of the visual word tree with size $S$. Therefore, we encode the visual word index $v$, and the number of bits occupied by this part is:
\begin{align}
C_{i,n}^{kf,BoW} = {\log _2}({S}),
\end{align}
To represent feature descriptors lossless, we further calculate the residual vector between the feature $i$ and the closest word $v$. In \cite{ref55}, For a descriptor of an ORB feature, the residual vector is computed by XOR. For a binary descriptor of length $D$ containing $h$ non-zero elements, the number of bits required to encode a residual vector is:
\begin{align}
C_{}^{res}(h,p_0) = -(D-h)\cdot {\log _2}({p_0}) - h{\log _2}(1 - {p_0}),
\end{align}
where $p_0$ is the probability that the binary residual vector element is 0. The number of bits needed to calculate the residual vector between the descriptor of feature $n$ and the closest word in the compressed image frame $i$ is:
\begin{align}
C_{i,n}^{kf,res} = C_{}^{res}(h_{i,n}^{kf},p_0^{kf}),
\end{align}
\subsection{Robot-Edge-Cloud Communication}
TCP is a reliable protocol that guarantees that data will not be lost or corrupted during transmission. TCP ensures data integrity and reliability by establishing, transferring, and releasing connections. In order to avoid the system communication packet loss, the communication module is based on the socket programming using TCP protocol, and adopts the message packet loss retransmission mechanism to communicate \cite{ref45}. In contrast to other methods, in socket programming the client creates a connection socket and sends requests to the server, so communication in the robot, the edge server, and the cloud allows dynamic joining and requesting communication during the task.

\subsubsection{Robot-Edge}
The communication between the robot and the edge is bidirectional. We adopt the efficient communication mode in \cite{ref45} to ensure that the message will not be sent repeatedly. At the same time, when the data packet loss is detected, data will be resold. The network bandwidth is reduced as much as possible through data compression and encoding of keyframes and non-keyframes, respectively. When the edge server receives the data, it decodes it and maintains the local VIO through the data.

The initialization of the IMU is placed on the VIO's local mapping thread, which allows for more precise inertial parameters. In order to reduce the influence of the inaccuracy of the off-line calibration parameters on the pre-integration accuracy of the robot IMU. After the initialization of the IMU, the edge server transmits better inertial parameters to the robot.
\subsubsection{Edge-Cloud}
In order to share the local map data from the edge to the cloud server and ensure the accuracy of the local odometry of the edge server, we establish the bidirectional communication of the Edge-Cloud.
The edge server will notify the cloud server of changes to the keyframe data and map information in its local map, and the entire data structure of the key frame and mappoints will be continuously sent (the keyframe data includes compressed data of 2D feature points extracted from the image), which accounts for the main traffic between the edge server and the cloud server. Each compressed 2D feature key takes up about 220 bytes. Other data that takes up less traffic include IMU related information (IMU measurements and current bias and velocity estimates) and confirmation messages for keyframe and map point message updates. After receiving these messages, the cloud server updates the keyframe pose and global map through global graph optimization. After the global image is optimized we pass the pose of the updated keyframe to the edge. This quantifies the drift in the current attitude estimation.
\section{VIO \& Global Map Fusion}
\subsection{Tracking and Local Mapping}
\begin{figure}[h]
	\centering
	\subfigure[Visual-Inertial joint optimization factor graph during initialization]{
		\includegraphics[width=0.45\textwidth]{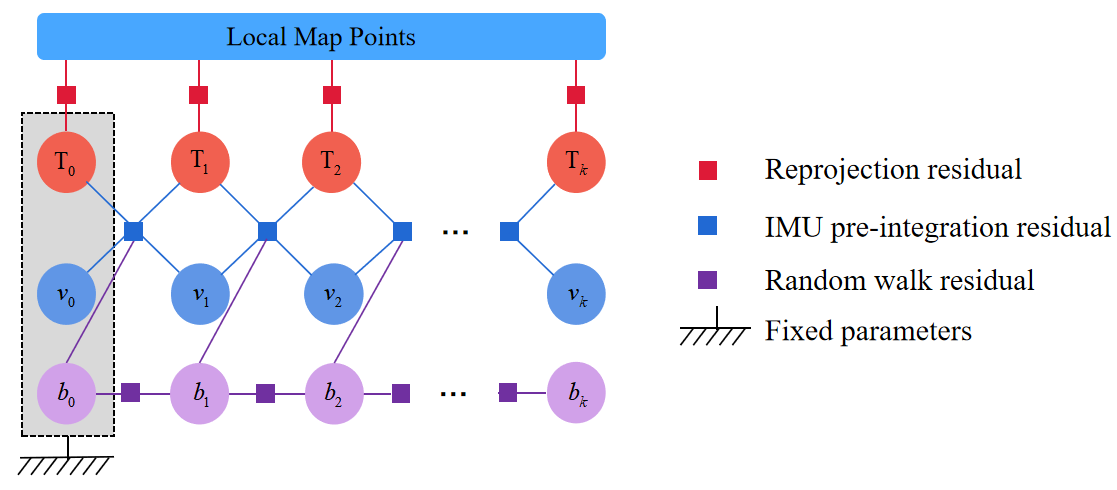}
	}\hspace{-12mm}
	\quad
	\subfigure[Visual-Inertial joint optimization factor graph in local mapping thread]{
		\includegraphics[width=0.45\textwidth]{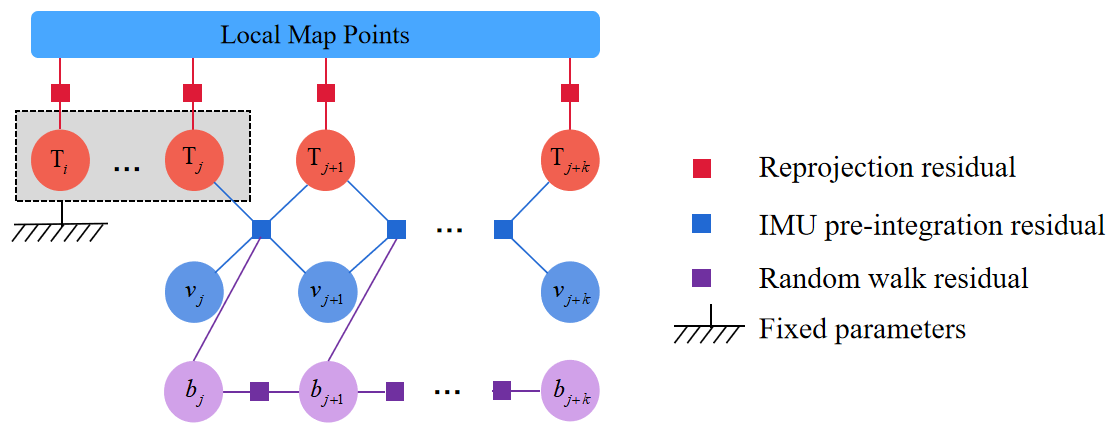}
	}
	\caption{Visual-Inertial joint optimization factor graph.}
	\label{FIG:5}
	\vspace*{-0cm}
\end{figure}

In the initialization stage, after Pure-Visual optimization estimation and Only-Inertial optimization estimation, the initial mappoints are accurately represented, while visual parameter information and inertial parameter information are accurately estimated, relatively. On this basis, the parameters are further refined by Visual-Inertial joint optimization as shown in Fig. \ref{FIG:5} (a), and the scale is rapidly converged as an explicit variable. It is worth noting that here we fixed the initialization of the pose and parameters of the keyframe and only optimized the scale and gravity information. For the tracking thread, the method described in Section \uppercase\expandafter{\romannumeral4} is used for inter-frame feature matching. A simple and fast sliding window optimization is used to optimize the pose of the latest frame, while the mappoints remain the same.

All keyframes participate in the local mapping thread. Then the proposed system transmit the generated mappoints, optimized keyframe pose and parameters to the cloud server through the communication module. At the same time, the IMU initialization process takes place in the local mapping thread. In this process, on the basis of manifold pre-integration in Equation (4), the rotation, velocity and position residual of the IMU between frame $i$ and frame $j$ can be expressed as:

\begin{align}
\begin{array}{l}
e_{imu}^{i,j} = [e_{\Delta {{\rm{R}}_{i,j}}}^{i,j},e_{\Delta {{\rm{v}}_{ij}}}^{i,j},e_{\Delta {{\rm{p}}_{ij}}}^{i,j}]\\
e_{\Delta {{\rm{R}}_{ij}}}^{i,j} = {\mathop{\rm Log}\nolimits} \left[ {\Delta {\rm{R}}_{ij}^ \top {\rm{R}}_i^ \top {{\rm{R}}_j}} \right]\\
e_{\Delta {{\rm{v}}_{ij}}}^{i,j} = {\rm{R}}_i^ \top ({{\rm{v}}_j} - {{\rm{v}}_i} - g_W\Delta {t_{ij}}) - \Delta {{\rm{v}}_{ij}} \\
e_{\Delta {{\rm{p}}_{ij}}}^{i,j} = {\rm{R}}_i^ \top ({{\rm{p}}_j} - {{\rm{p}}_i} - {{\rm{v}}_i}\Delta {t_{ij}} - \frac{1}{2}g_W\Delta {t_{ij}}^2) - \Delta {{\rm{p}}_{ij}}
\end{array},
\end{align}
where ${\rm{R}}_{ij}$, ${\rm{v}}_{ij}$ and ${\rm{p}}_{ij}$ are the results of pre-integration of IMU after bias update, respectively. Their calculation process can be expressed as:
\begin{align}
\begin{array}{l}
\Delta {\rm{R}}_{ij}^{} = \Delta {{{\rm{\tilde R}}}_{ij}} \cdot {\mathop{\rm Exp}\nolimits} \left( {\frac{{\partial \Delta {{{\rm{\tilde R}}}_{ij}}}}{{\partial {\rm{b}}_i^g}}} \right)\\
\Delta {{\rm{v}}_{ij}} = \Delta {{{\rm{\tilde v}}}_{ij}} + \frac{{\partial {{{\rm{\tilde v}}}_{ij}}}}{{\partial {\rm{b}}_i^g}}\partial {\rm{b}}_i^g + \frac{{\partial {{{\rm{\tilde v}}}_{ij}}}}{{\partial {\rm{b}}_i^a}}\partial {\rm{b}}_i^a\\
\Delta {{\rm{p}}_{ij}} = \Delta {{{\rm{\tilde p}}}_{ij}} + \frac{{\partial {{{\rm{\tilde p}}}_{ij}}}}{{\partial {\rm{b}}_i^g}}\partial {\rm{b}}_i^g + \frac{{\partial {{{\rm{\tilde p}}}_{ij}}}}{{\partial {\rm{b}}_i^a}}\partial {\rm{b}}_i^a
\end{array}.
\end{align}
The reprojection error of mappoint $l$ in frame $k$ can be expressed as:
\begin{align}
{e_{cam}^{l,k}} = {{\rm{x}}^k} - {\Pi _{cam}}({{\rm{R}}_k}{{\rm{X}}^l} + {{\rm{t}}_k})
\end{align}
where ${{\rm{\Pi}}_{cam}} {\left[ {\begin{array}{*{20}{c}}
		X\\
		Y\\
		Z
		\end{array}} \right]} {\rm{ = }}\left[ {\begin{array}{*{20}{c}}
	{{f_x}\frac{X}{Z} + {c_x}}\\
	{{f_y}\frac{X}{Z} + {c_y}}
	\end{array}} \right]$ is the projection function of the camera, $  {\rm X}^l \in {\mathbb R}^3$ 9s the 3D coordinates of mappoint $l$ in the world coordinate, and $\rm x_k \in {\mathbb R}^2$ is the observation point for ${\rm{X}}^l$ on frame $k$. As shown in Fig. \ref{FIG:5} (b), the corresponding node of the Local BA optimization problem can be expressed as:
\begin{align}
\begin{array}{l}
\chi  = [{\chi _0},{\chi _1},{\chi _2}, \cdots ,{\chi _k}],\\
{\cal L} = [{{{\cal L}}_0},{{{\cal L}}_1},{{{\cal L}}_2}, \cdots ,{{{\cal L}}_l}],
\end{array}
\end{align}
where ${\chi _i} = \{\rm R_i, \rm t_i,\rm v_i, \rm b_i^g, \rm b_i^a \}$ represents the state vector corresponding to the keyframe. $\rm b_i^g$ and $\rm b_i^a$ are gyroscope bias and accelerometer bias at time $i$, respectively. The random walk errors programmed to connect adjacent moments can be modeled by Wiener process. ${{\cal L}}_i$ is the coordinate set of covisible mappoints.

The edges in the corresponding Local BA optimization problem include reprojection residuals of covisible mappoints is expressed as $e_{cam}$, and IMU pre-integral residual between keyframes is expressed as $e_{imu}$. So the whole loss function can be expressed as:
\begin{align}
\mathop {\min }\limits_{\chi ,{\cal L}} \{ \sum\limits_{j \in C} \rho ({{{\left\| {e_{imu}^{j,j + 1}} \right\|}^2}})  + \sum\limits_{l \in L}\sum\limits_{k \in C} {\delta ({{\left\| {e_{cam}^{l,k}} \right\|}^2}} )\} 
\end{align}
where $C$ represents the set of newly added keyframes in the Local BA window, and $L$ represents the mappoints with covisible relationship between keyframes. To keep the results robust, we add Cauchy loss function ${\delta ( \cdot )}$ and Huber loss function ${\rho ( \cdot )}$ to improve the robustness of the loss function. 

Local mapping is performed after new keyframes are generated, but the optimization of all keyframes solving Eq. (18) is computationally intensive. As shown in Fig. 5 (b), during the mapping process, we fix the pose of the older frame unchanged, and only optimize the mappoints seen by the keyframes in a fixed local window.
\subsection{Map Fusion}
\begin{figure}[h]
	\centering
	\includegraphics[width=0.45\textwidth]{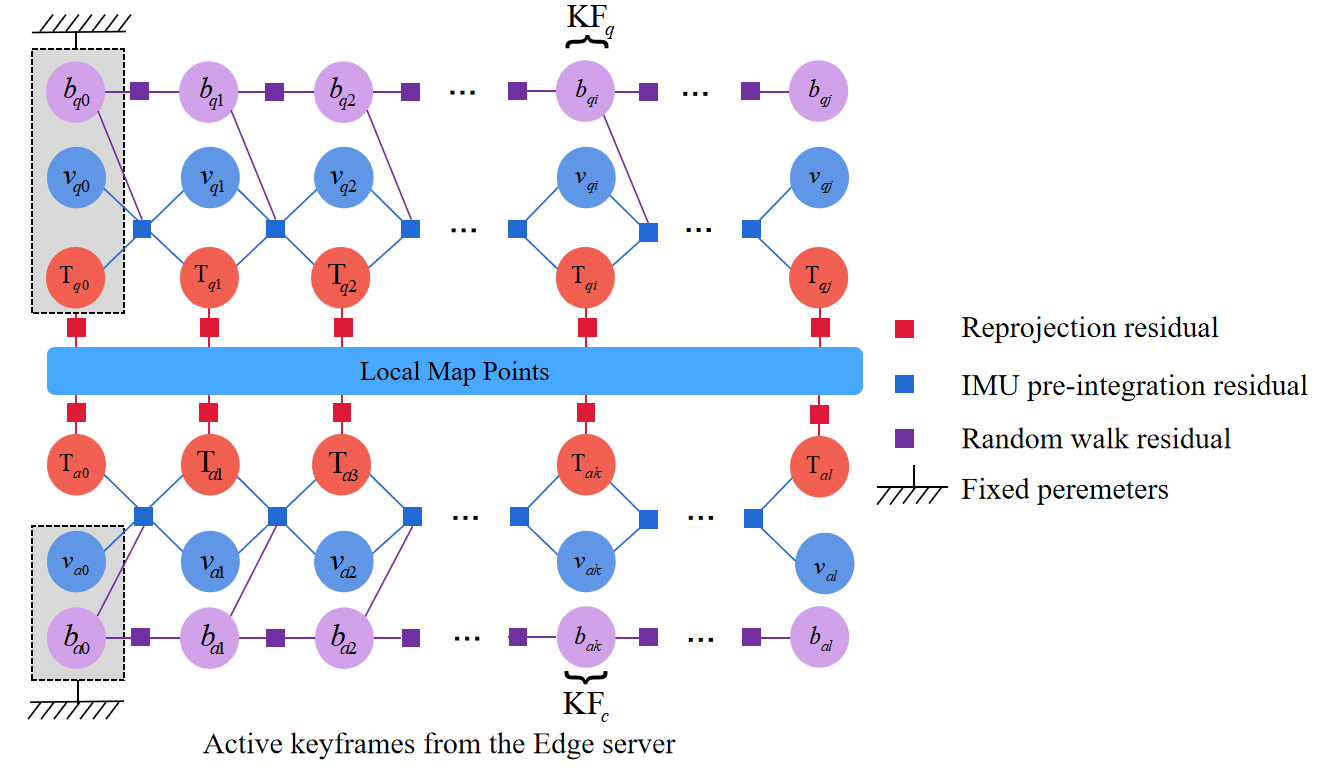}
	\caption{Visual-Inertial joint optimization factor graph in the cloud.}
	\label{FIG:6}
	\vspace*{-0.25cm}
	\centering
\end{figure}

After local mapping is done from the edge, Global Pose-Graph Optimization (GPGO) occurs in the cloud. In our method, we do not distinguish between the closed loop between the inter-robot or the intra-robots. As shown in Fig. \ref{FIG:6}, we used the DBoW method to check candidate keyframes. When candidate keyframe $\rm {KF}_c$ was successfully detected, the relative pose $\rm{T}_{qc}$ between co-view keyframe $\rm {KF}_q$ and $\rm {KF}_c$ is optimized by feature matching and refining reprojection errors. When  $\rm {KF}_q$ and  $\rm {KF}_c$ are from the same robot, closed-loop optimization is carried out inside the intra-robot. When $\rm {KF}_q$ and $\rm {KF}_c$ are from different robots, the relative position between inter-robot is optimized. Although the pose accuracy after optimization will be higher with more keyframes, the number of keyframes will be huge with the increase of the number of robots, which will also greatly increase the computing burden. According to the method in \cite{ref47}, after eliminating redundant keyframes, the cloud GPGO problem can be described as:
\begin{align}
\mathop {\min }\limits_{\chi ,{\cal L}} \{  {{\left\| {e_{prio}^c} \right\|}^2} +  \sum\limits_{j \in C} \rho ({{{\left\| {e_{imu}^{j,j + 1}} \right\|}^2}})  + \sum\limits_{l \in L}\sum\limits_{k \in C} {\delta ({{\left\| {e_{cam}^{l,k}} \right\|}^2}} )\} 
\end{align}
where $e_{prio}^c$ is the pose constraint that fixes the first keyframe. Different from local optimization, GPGO occurs on all keyframes and mappoints, while constraints imposed by IMU measurement only occur between successive keyframes of the same robot.

Although more keyframes and mappoints can improve the accuracy of optimization results, with the enlargement of the scene, more and more keyframes and mappoints will increase the calculation burden of GPGO. In the \cite{ref47}, they do pruning by calculating the similarity of keyframes and observations of mappoints. But this sometimes cuts out well-observed mappoints. At the same time, greedy removal of keyframes will affect the accuracy of the final optimization. Inspired by \cite{add4}, we reduce the storage of the map and ensure the GPGO speed by MBP (extracting the backbone of the map and creating virtual key frames through good observations). 
\section{Experiments}
We evaluate our system on a publicly available EuRoC dataset containing three scenarios: a machine hall (MH\_01-MH\_05 sequence), indoor room 1 (V1\_01-V1\_03 sequence), and indoor room 2 (V2\_01-V2\_03 sequence). In this dataset, camera data is captured by Aptina MT9V034 global shutter, IMU data is captured by ADIS16448, and groundtruth is captured by Leica MS50 laser tracker. The internal parameters of the camera and the the external parameters of the camera-IMU are obtained by the Kalibr \cite{ref60}. The algorithm is tested on real devices. Our server and mobile robot platforms are equipped with models of:
\begin{itemize}
	\item robot: Intel(R) Core(TM) i3-9100 CPU 8GB RAM
	\item Edge Serve: Intel(R) Core(TM) i7-9750H CPU 8GB RAM.
	\item Cloud Serve: Amazon Web Services (AWS) cloud serve.
\end{itemize}
In addition, we run the proposed algorithm in a real scene. The details are covered in Section \uppercase\expandafter{\romannumeral7}-D.
\subsection{Optical Flow Tracking Accuracy and Speed}
We first tested IMU-assisted non-keyframes feature matching. We use the method in Section \uppercase\expandafter{\romannumeral4} to calculate matches between non-keyframes in a way that does not require calculating descriptors. The front-end of our method is improved based on the ORB-SLAM3 algorithm and the LK algorithm. 

\begin{figure*}[htpb]
	\centering
	\subfigure[Raw images]{
		\includegraphics[width=0.3\textwidth]{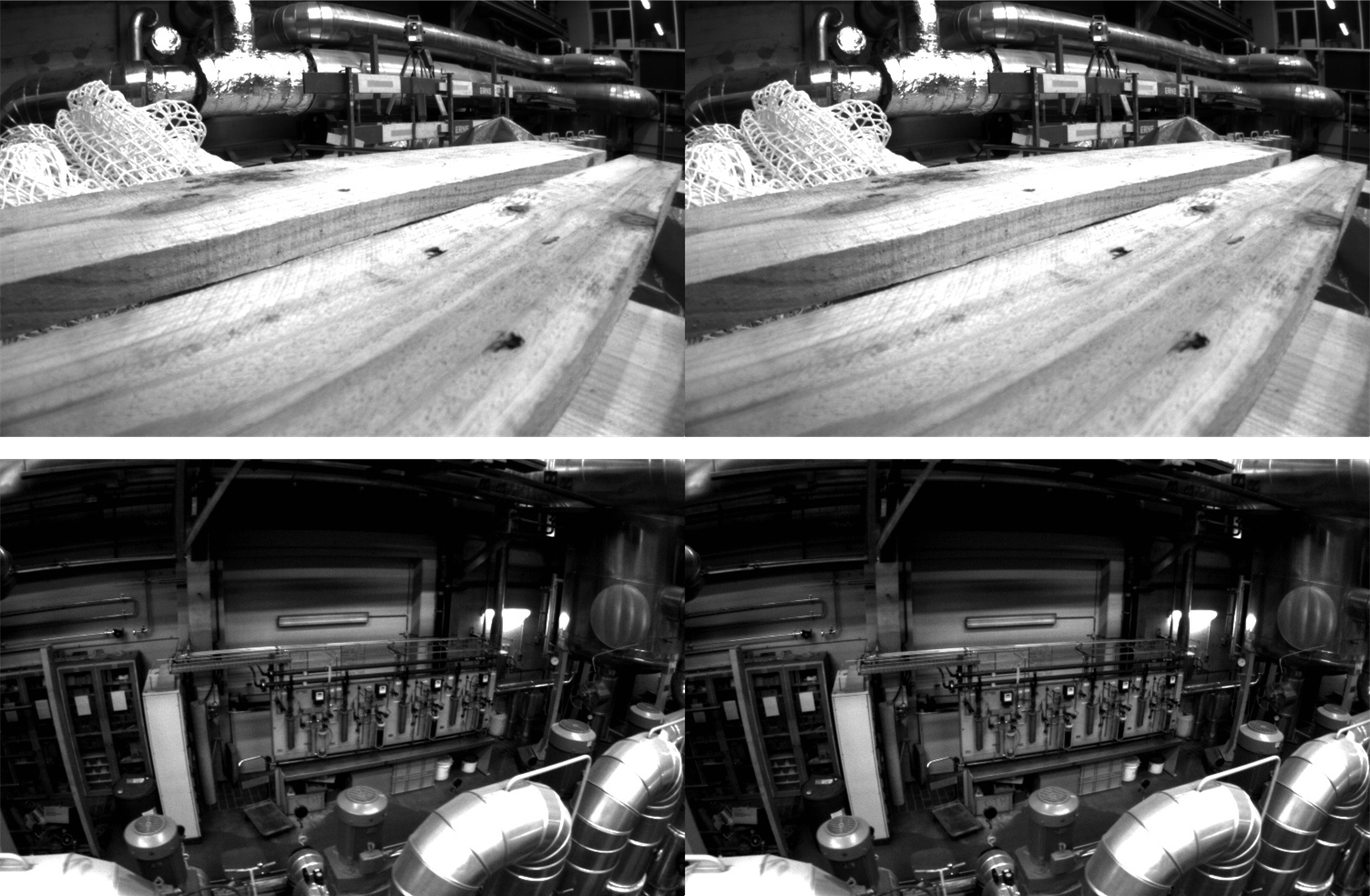}
	}
	\subfigure[LK w/o IMU preintegration]{
		\includegraphics[width=0.3\textwidth]{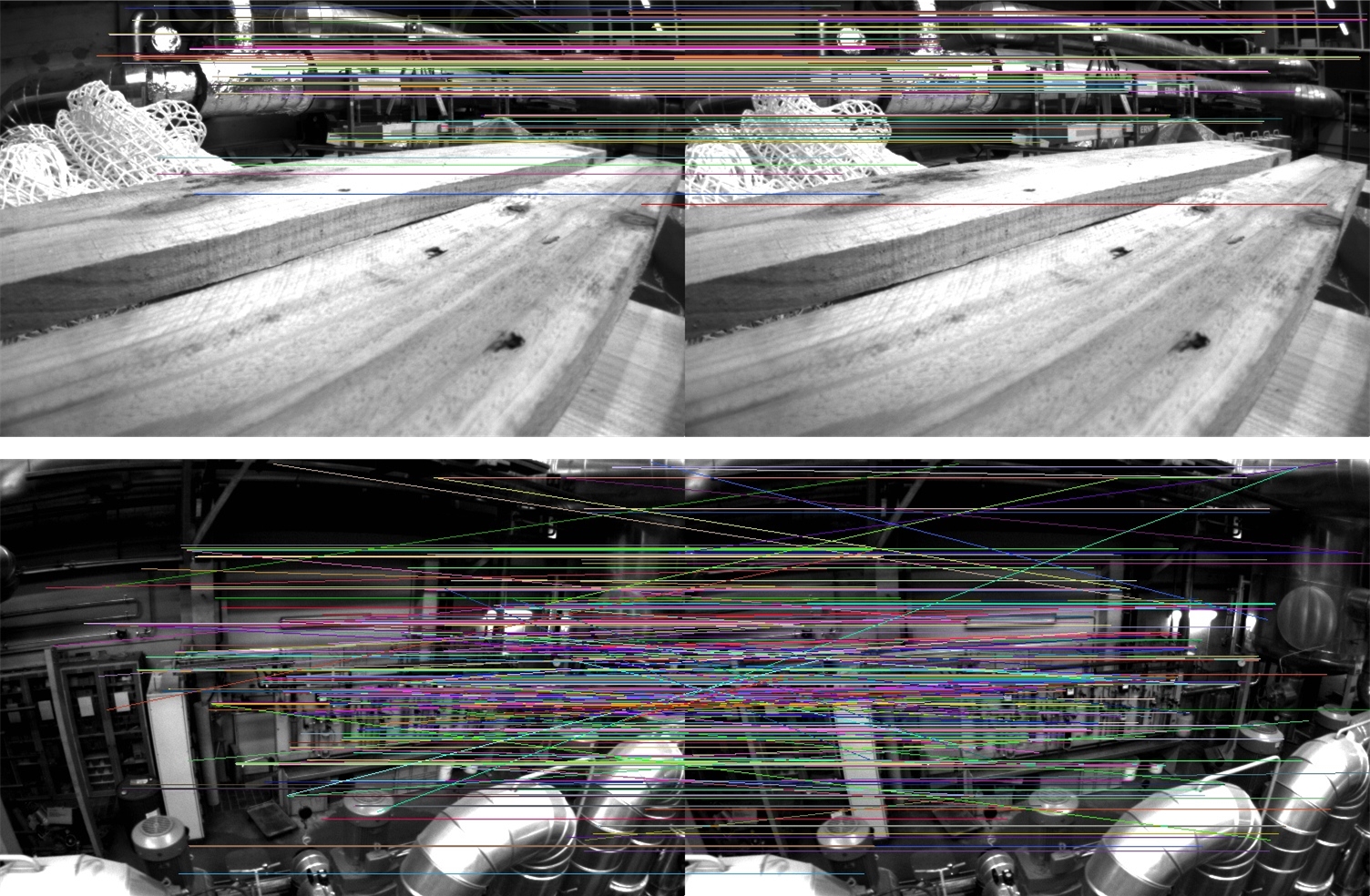}
	}
	\subfigure[LK w/ IMU preintegration (The proposed)]{
		\includegraphics[width=0.3\textwidth]{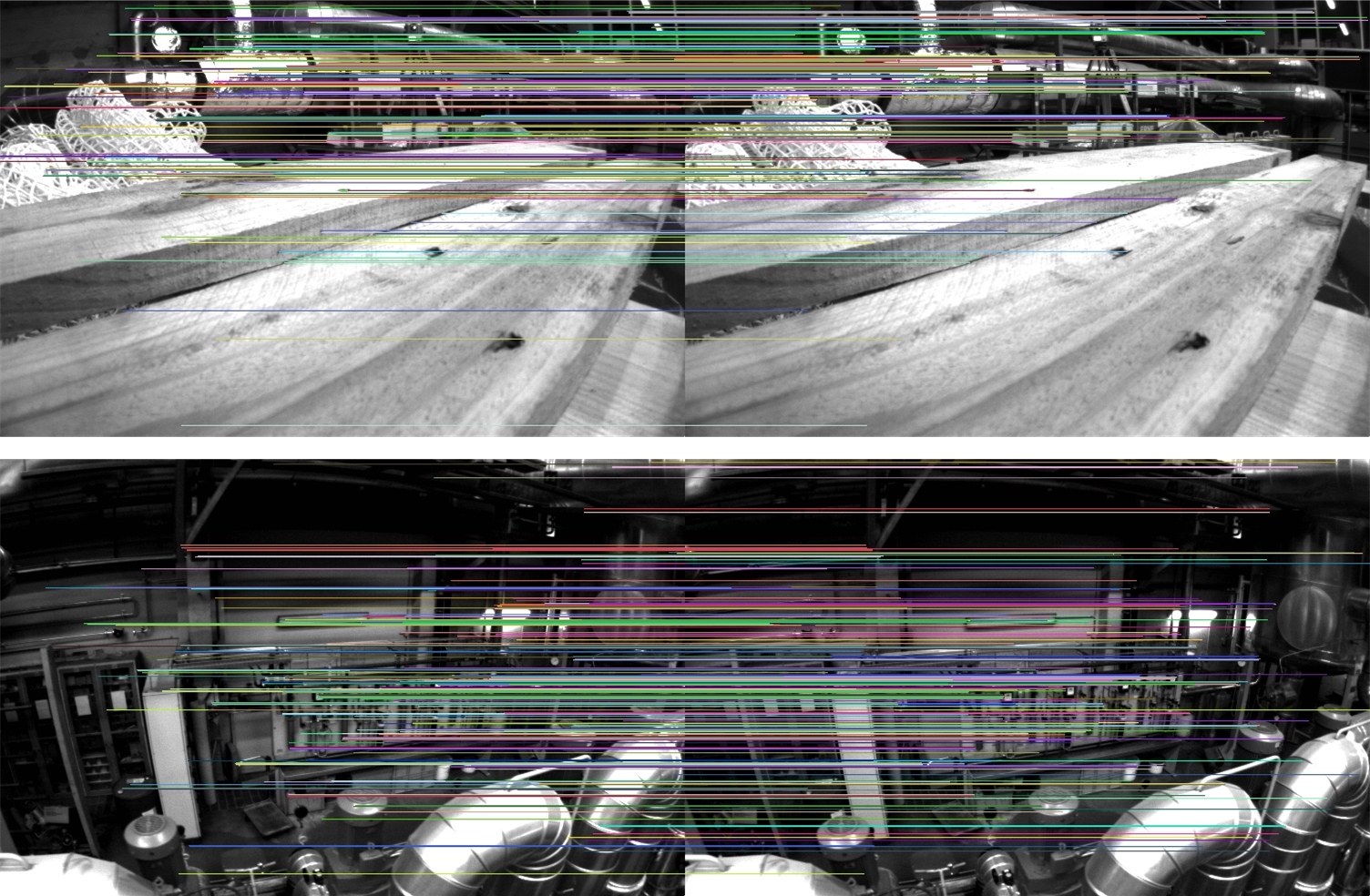}
	}
	\caption{The tracking results of traditional LK optical flow tracking and IMU-assisted optical flow tracking on EuRoC. Raw image data (a), traditional LK optical flow tracking results without IMU participation (b), IMU-assisted LK optical flow tracking results (c).}
	\label{FIG:7}
	\vspace*{-0.45cm}
\end{figure*}

\begin{figure*}[htbp]
	\centering
	\includegraphics[width=0.75\textwidth]{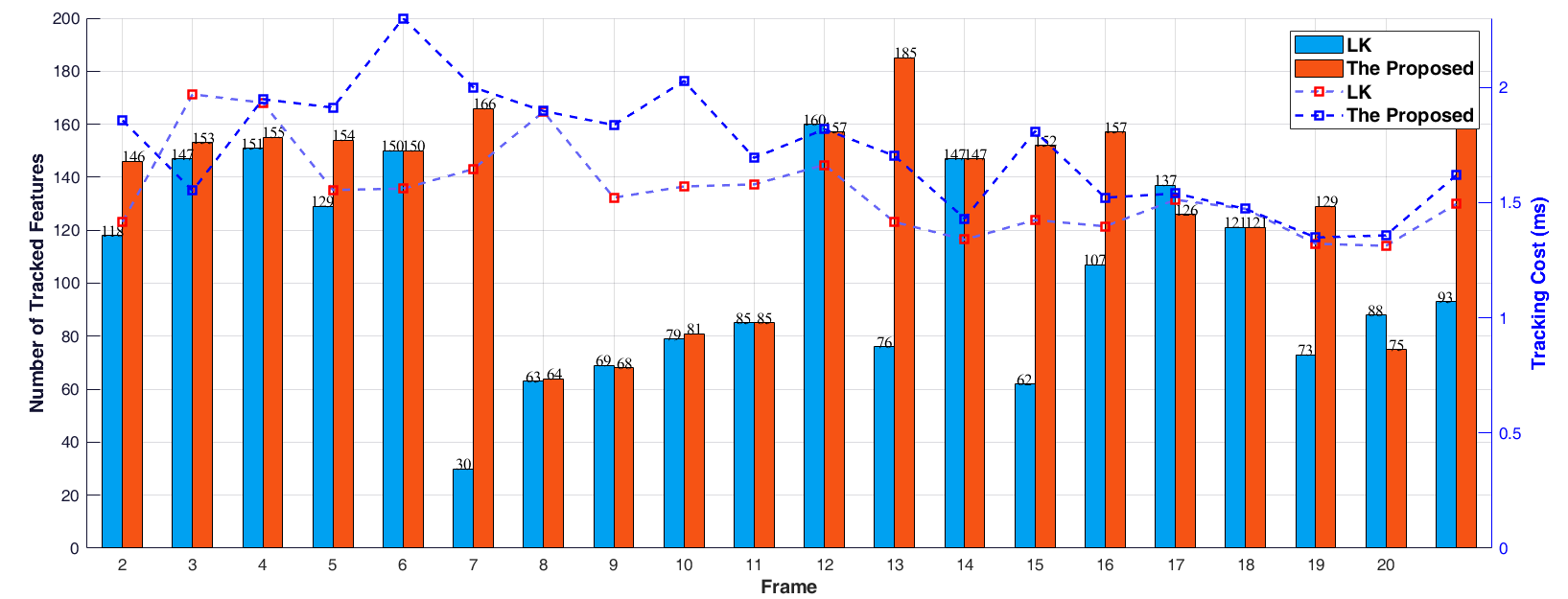}
	\caption{The number of points tracked and the track time spent. The blue bar represents the number of points tracked by LK optical flow method, and the orange bar represents the number tracked by the proposed. The purple broken line represents the time spent in LK tracking and the green broken line represents the time spent in the proposed tracking.}
	\label{FIG:8}
	\centering
\end{figure*}

Fig. \ref{FIG:7} shows our tracking results on EuRoC using traditional LK optical flow tracking and IMU-assisted optical flow tracking. In the results, lines of different colors connect the two matched feature points. It can be seen that compared with the traditional LK optical flow tracking method, the proposed tracking method can match more points and reduce the number of mismatched points. In addition, in Fig. \ref{FIG:8}, we compare the tracking speed and the number of points tracked by the two methods for 20 consecutive frames in the EuRoC dataset. In some frames (e.g. 8 to 11 frames), the gain from IMU pre-integration is limited due to significant changes in illumination. In addition, because RANSAC's random selection process makes it unstable, we cannot guarantee a stable gain in all successive frames. It can be seen from the results that in most cases, compared with LK optical flow method, our method can significantly improve the number of tracking points and tracking speed. This is mainly because IMU pre-integration can improve a good initial value for the feature points to be matched. Therefore, stable tracking provided by IMU-assisted non-keyframes feature matching can improve the stability of localization.

\subsection{Communication Rate}
\begin{figure}[htbp]
	\centering
	\includegraphics[width=0.35\textwidth]{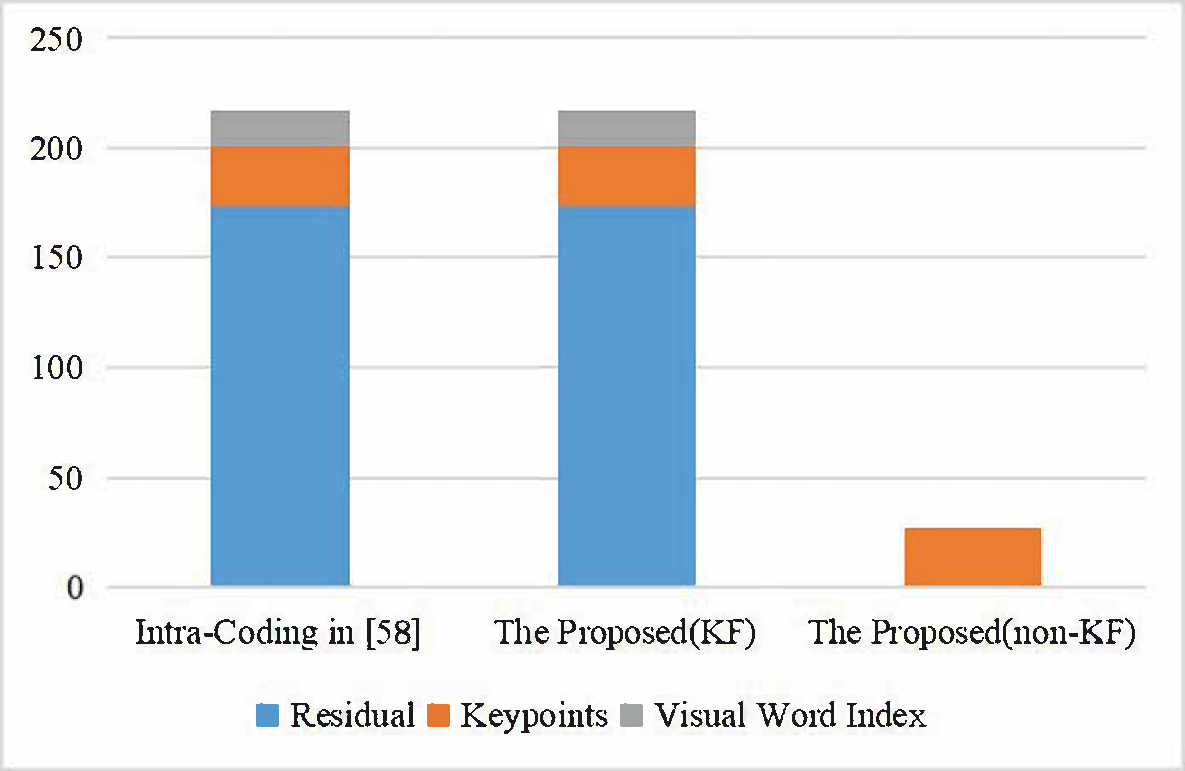}
	\caption{The proportion of each part of each feature in different methods under lossless compression.}
	\label{FIG:9}
	\centering
\end{figure}

In this cloud-edge-user system, we focus on achieving high localization accuracy through low latency and low bandwidth. Therefore, we compare with the current most advanced algorithms in the aspects of client communication delay and bandwidth occupancy. Compared with the most advanced local binary feature compression methods \cite{ref55}, our method classifies keyframes and non-keyframes for transmission. Non-key frame matching based on optical flow is accomplished without descriptors. Table \ref{TAB:1} shows the size of memory required for the proposed and \cite{ref55} to transmit a feature in the EuRoC dataset. When transferring keyframes, our method uses the same 217 bits of memory as the intra-frame encoding mode in \cite{ref55} when transferring a feature. In non-keyframes, our method greatly reduces the amount of data because it does not need to transmit descriptors and visual word index. The proportions of feature points, visual word indexes and residual vectors when transmitting a feature are shown in detail in Fig. \ref{FIG:9}. In addition, we conduct a separate data transmission experiment on EuRoC. We calculate the average of the communication rate from the robot to the edge server across 8 sequences in the EuRoC. The communication rate by the method in \cite{ref55} is 211.9kbits/\textit{s}, and the proposed method is 46.81kbits/\textit{s}.
\begin{table}[]
	\centering
	\caption{Comparison of the memory required for a feature point}
	\begin{tabular}{c|l|l|l}
		\hline
		\multirow{2}{*}{Methods}  & \multirow{2}{*}{Intra-Coding in \cite{ref55}} & \multicolumn{2}{c}{The Proposed} \\ 
		\cline{3-4} 
		& &KF        & Non-KF     \\ \hline
		\multicolumn{1}{l|}{Memory} & 217 bits         & 217 bits          & 27 bits    \\ \hline
	\end{tabular}
    \label{TAB:1}
\end{table}

\subsection{Localization and Map Fusion Accuracy}
For a centralized multi-robot SLAM system, the main indexes of evaluation are global consistency and trajectory accuracy. Therefore, the root-mean-square error (RMSE) of the absolute trajectory error (ATE) is used to evaluate the collaborative positioning accuracy of MH\_01-MH\_03, MH\_01-MH\_05, V1\_01-V1\_03, and V2\_01-V2\_03, respectively. The RMSE of ATE and can be
given as:
\begin{align}
{\rm{AT}}{{\rm{E}}_{{\rm{RMSE}}}} = \sqrt {\frac{1}{N}(\sum\limits_{i = 1}^N {{{\left\| {trans({\rm{T}}_{\rm{G}}^{ - 1}{{\rm{T}}_{\rm{E}}})} \right\|}^2}} )}
\end{align}
where ${\rm{T}}_{\rm{G}}$ and ${\rm{T}}_{\rm{E}}$ represent the pose of groundtruth trajectory and estimated trajectory, respectively. $N$ represent is the number of frames the trajectory contains. The ATE is the direct difference between the estimated pose and the real pose, which can directly reflect the algorithm accuracy and the global trajectory consistency.

We run our architecture on the robot side (i3 CPU), edge serve (i7 CPU), and AWS cloud server. Table \ref{TAB:2} shows a quantitative comparison of the ATE RMSE obtained by CVIDS with CCM-SLAM \cite{ref44}, VINS-mono \cite{ref26}, ORB-SLAM3 \cite{ref8}, CVIDS \cite{ref48} and COVINS AWS \cite{ref47}.
Due to the different system architectures, we obtain the average ATE RMSE for multi-robot localization on multiple sequences from the existing references. It can be seen from the results that the average RMSE of our method in all sequences is consistent with the current most advanced centralized method, and even better than most results, which confirms the superiority of its positioning. In addition, the proposed with MBP performed better in all four groups of experiments than the proposed without MBP. This shows that MBP can bring some gains to the edge-assisted multi-robot SLAM. It is worth mentioning that in the experiment, we utilized the multi-session mode of VINS-Mono and ORB-SLAM3. These data all confirm the improvements our architecture brings to multi-robot systems. However, in the MH\_01-MH\_05 sequence, the proposed achieves poor localization with COVINS. However, according to our verification \cite{r3}, the speed improvement of the front end based on optical flow method is obvious compared with the front end based on feature point method in COVINS. Further, to demonstrate the advantages of MBP in memory and running speed, we compare the specific indicators of MH\_01-MH\_05 sequences between the proposed with MBP and the proposed without MBP. The results show that the proposed with MBP can guarantee high accuracy by optimizing fewer keyframes and mappoints.
\begin{table*}[!htbp]
    \renewcommand{\arraystretch}{1.1}
	\centering
	\caption{RMSE of ATE in units of on the EuRoC dataset. Our method is to calculate the average after five runs. Note that the bold means the best, and "-" means that the result of the corresponding method cannot be found in the literature.}
   \begin{tabular}{|c|c|c|c|c|c|}
   	\hline
   	
   	\multicolumn{2}{|c|}{ \multirow{2}*{} }& \multicolumn{4}{c|}{Sequences}\\
   	\cline{3-6}
   	\multicolumn{2}{|c|}{}&MH\_01-MH\_03&MH\_01-MH\_05&V1\_01-V1\_03&V2\_01-V2\_03\\
   	\hline
   	\multirow{6}*{Method}&CCM-SLAM \cite{ref44}&0.077&-&-&-\\
   	\cline{2-6}
   	&VINS-mono \cite{ref26}&-&0.210&-&-\\
   	\cline{2-6}
   	&ORB-SLAM3 \cite{ref8}&0.037&0.065&0.040&0.048\\
   	\cline{2-6}
   	&CVIDS \cite{ref48}&-&0.074&-&-\\
   	\cline{2-6}
   	&COVINS AWS \cite{ref47}&\textbf{0.025}&\textbf{0.039}&0.049&-\\
   	\cline{2-6}
   	&The Proposed (w/o MBP)&0.028&0.051&0.035&0.045\\
   	\cline{2-6}
   	&The Proposed (w/ MBP)&\textbf{0.025}&0.046&\textbf{0.031}&\textbf{0.033}\\
   	\hline
   \end{tabular}
   \label{TAB:2}
\end{table*}

\begin{table*}[!htbp]
	\renewcommand{\arraystretch}{1.1}
	\centering
	\caption{Comparison of results using different redundant keyframe removal methods.}
	\begin{tabular}{|c|c|c|c|}
		\hline
		Methods &ATE/m&Number of Keyframes&Number of Mappoints\\
		\hline
		The Proposed (w/o MBP)&0.051&1550&57169\\
		\hline
		The Proposed (w/ MBP)&\textbf{0.046}&\textbf{925}&\textbf{41923}\\
		\hline
	\end{tabular}
	\label{TAB:3}
\end{table*}

\begin{figure}[h]
	\centering
	\subfigure[Map and covisible relationships]{
		\includegraphics[width=0.33\textwidth]{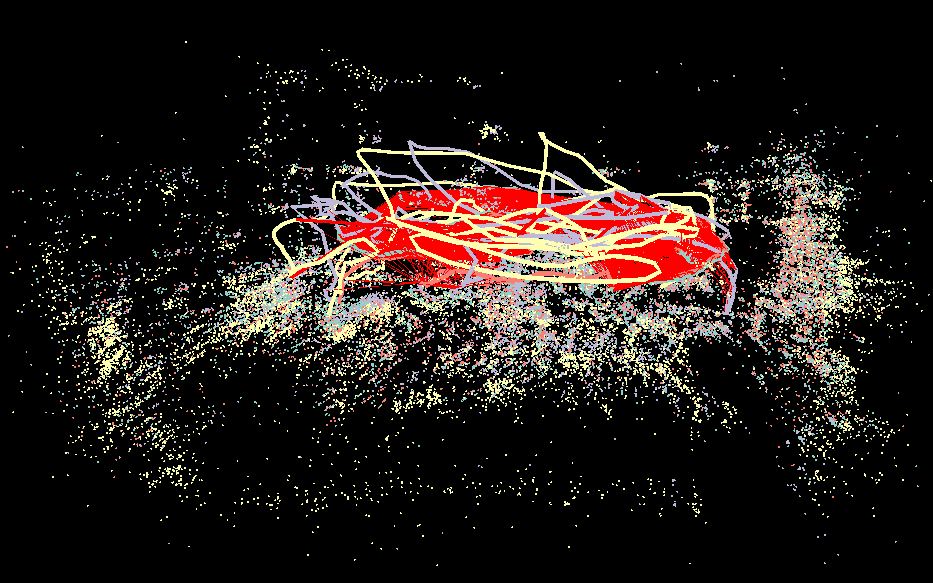}
	}\hspace{-6mm}
	\quad
	\subfigure[Trajectories and closed-loop keyframes]{
		\includegraphics[width=0.33\textwidth]{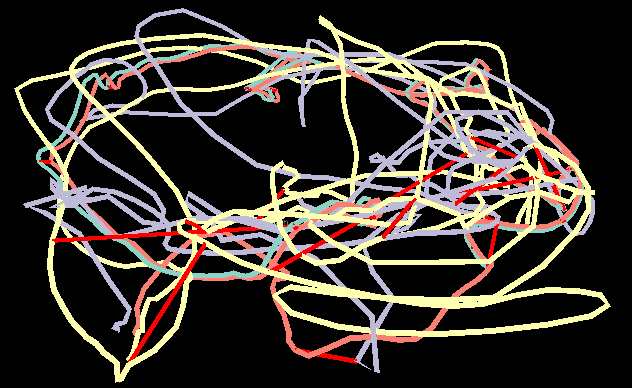}
	}
	\caption{The results of collaborative localization and mapping in the V2 sequences of the proposed method.}
	\label{FIG:10}
\end{figure}

\begin{figure}[h]
	\centering
	\subfigure[V2\_01]{
		\includegraphics[width=0.235\textwidth]{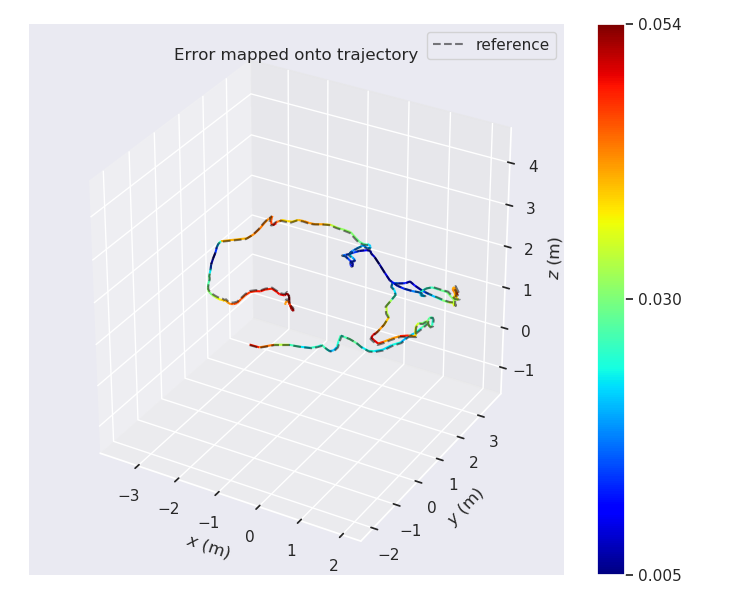}
	}\hspace{-4mm}
	\subfigure[V2\_02]{
		\includegraphics[width=0.235\textwidth]{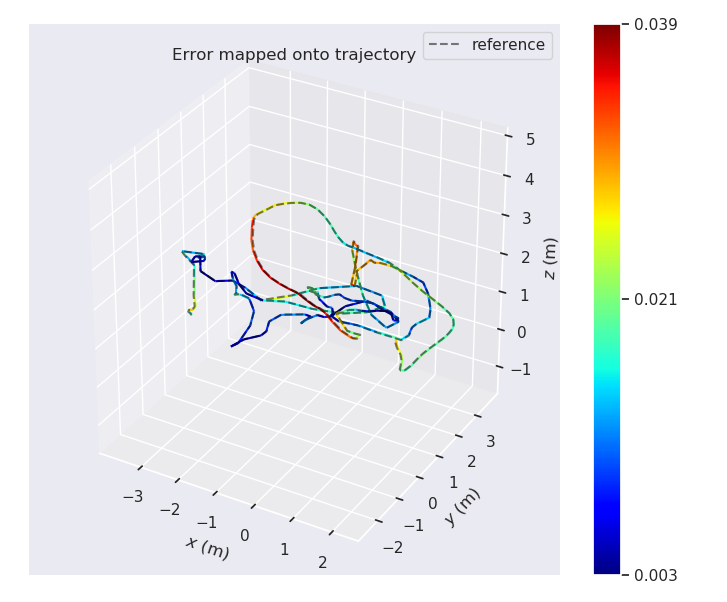}
	}\hspace{-4mm}
    \subfigure[V2\_03]{
    \includegraphics[width=0.235\textwidth]{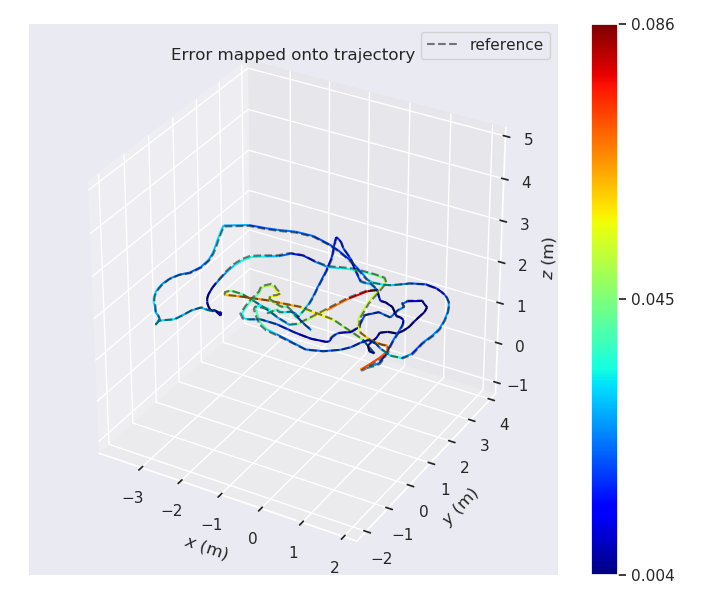}
    }\hspace{-4mm}
    \subfigure[V2\_01-V2\_03]{
    	\includegraphics[width=0.235\textwidth]{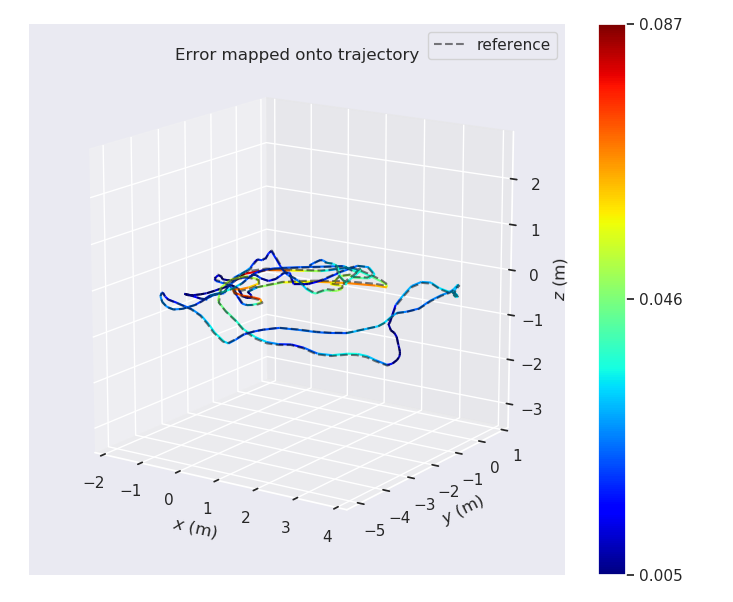}
    }
	\caption{The trajectories of collaborative localization and mapping in the V2 sequences of the proposed method.}
	\label{FIG:11}
\end{figure}

We map the sequences in the V2 scenario that was challenging with optical flow. Fig. \ref{FIG:10} shows the results of our collaborative SLAM in three sequences. Different colors represent different sequences of tracks and map points. The red connecting lines in Fig. \ref{FIG:10} (a) represent the common view, and the red connecting lines in Fig. \ref{FIG:10} (b) represent the closed-loop keyframes between the robots. To show the robot trajectories of the three sequences in the V2 scene clearly , we use the evo tool to align the generated trajectories with the reference ground truths. As shown in Fig. 11, we drew separate trajectories and collaborative localization trajectory of V2\_01, V2\_02 and V2\_03, respectively. It can be seen from the results that the trajectory of each robot and the combined trajectory can be well aligned with the reference groundtruths.
\subsection{Real scene experiment}
\begin{figure}[h]
	\centering
	\includegraphics[width=0.475\textwidth]{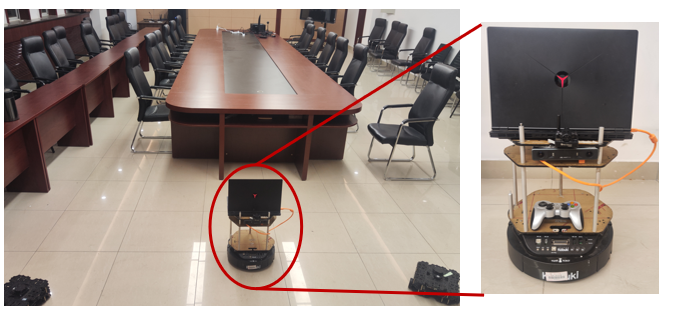}
	\caption{The experimental scene and the equipments.}
	\label{FIG:12}
	\centering
\end{figure}

We run our method in the indoor scene shown in Fig. \ref{FIG:12} using two turtlebot 3 (with Raspberry Pi 3) and turtlebot 2 (with Lenovo Y7000). The robot is equipped with the MYNT EYE Depth camera, and the joint calibration of the camera and IMU is carried out by kalibr. The groundtruth of the robots is obtained by Marvelmind\footnote{https://marvelmind.com/pics/architectures\_comparison.pdf} indoor "GPS" and IMU together. The robots circle for two weeks at the starting point and meet at the end of the trajectories. Throughout the process, we used the Lenovo Y7000 as the edge serve and AWS as the cloud server. In the real scene experiment, the internal parameters of the camera and the external parameters of the camera-IMU are calculated by Kalibr. We use three different robots to travel two weeks from one end of the room through different trajectories and meet through different paths to the end point without co-view.

\begin{figure}[h]
	\centering
	\subfigure[Map and Trajectories]{
		\includegraphics[width=0.222\textwidth]{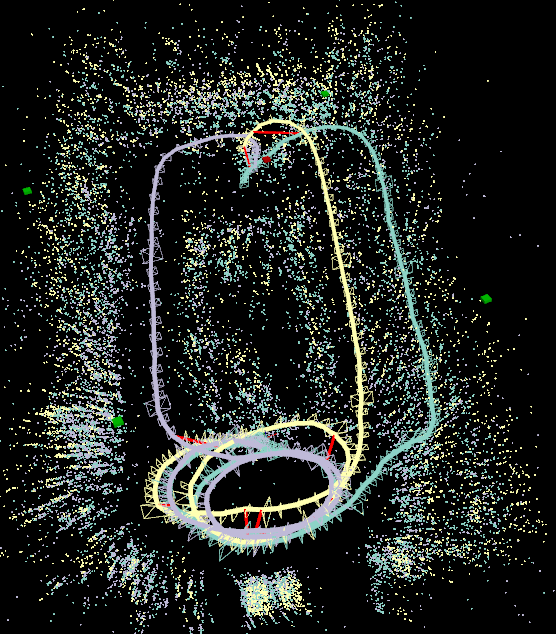}
	}\hspace{-4mm}
	\quad
	\subfigure[Trajectories and closed-loop keyframes]{
		\includegraphics[width=0.212\textwidth]{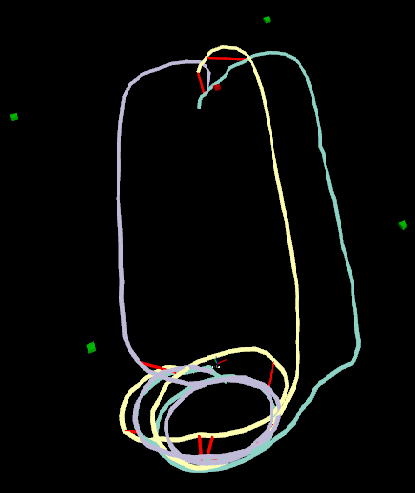}
	}
	\caption{The results of collaborative localization and mapping in the real scene of the proposed method.}
	\label{FIG:13}
\end{figure}

\begin{figure}[h]
	\centering
	\includegraphics[width=0.34\textwidth]{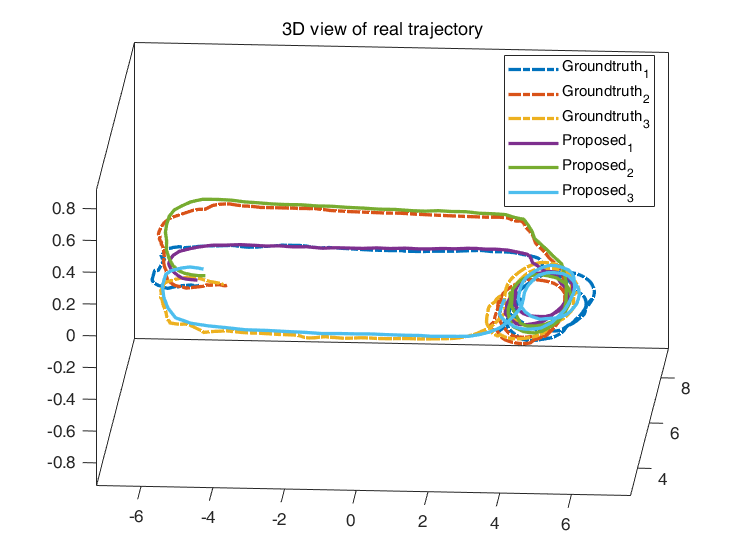}
	\caption{Comparison of measured trajectory and groundtruth in real scene.}
	\label{FIG:14}
	\vspace*{-0.25cm}
	\centering
\end{figure}

\begin{figure}[h]
	\centering
	\includegraphics[width=0.3\textwidth]{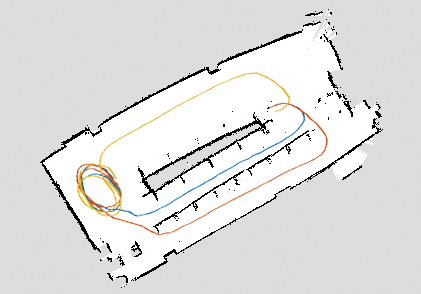}
	\caption{Trajectory overlaid with 2D grid map of the real scene for visual comparison.}
	\label{FIG:15}
	\centering
\end{figure}

The trajectories of different robots and the map points constructed are shown in Fig. \ref{FIG:13}. When measuring groundtruth via Marvelmind indoor "GPS", we used four fixed anchors (green squares in Fig. \ref{FIG:13} (b)) to locate the moving anchors (red squares in Fig. \ref{FIG:13} (b)). Further, by fusing IMU information, the reference groundtruth is calculated. The three robots loop at the starting point for two weeks and then form a closed loop at the end through different trajectories. It can be seen from the results that an obvious covisible relationship can be formed between the starting point and the end point. In Fig. \ref{FIG:14}, we align the estimated trajectory obtained by the proposed method with the reference groundtruth. For a better visual comparison, we align the estimated trajectory with the 2D grid map of the real scene in Fig. \ref{FIG:15}. The results show that our method has a good estimation effect in the actual scene.


\section{Conclusions}
In this paper, we propose a centralized multi-robot SLAM system for cloud, edge and client collaborative computing. The system can establish a global sparse map with the collaboration of multiple independent robots and estimate their positions in the global map. Through the research, we find that in the calculation and transmission process, the non-key frame descriptor is not necessary. However, the computation and transmission of descriptors consume a lot of computation, memory and transmission bandwidth. We build an IMU-assisted matching model between adjacent frames without descriptors. At the same time we prune and compress the transmitted data. Through experimental analysis, our matching method can achieve higher feature tracking accuracy and faster feature tracking speed. Meanwhile, we can complete collaborative SLAM with the least amount of data transmission compared with similar systems without loss of accuracy. After the global map fusion is completed in the cloud, the experimental comparison proves that our proposed method is competitive with the current most advanced multi-map system in terms of accuracy, speed and data transmission volume. In future work, we will solve the limitations of vision perception in complex scenes by integrating LiDAR. In addition, to avoid interference of our system such as dynamic environment, energy and battery limitations, sensor failures, in the future, more combinations of sensors and recognition methods based on deep learning will be added to our system.

%
\bibliographystyle{IEEEtran}
\bibliography{cas-refs.bib}
\vspace*{-1.2cm}
\begin{IEEEbiography}[{\includegraphics[width=1in,height=1.25in,clip,keepaspectratio]{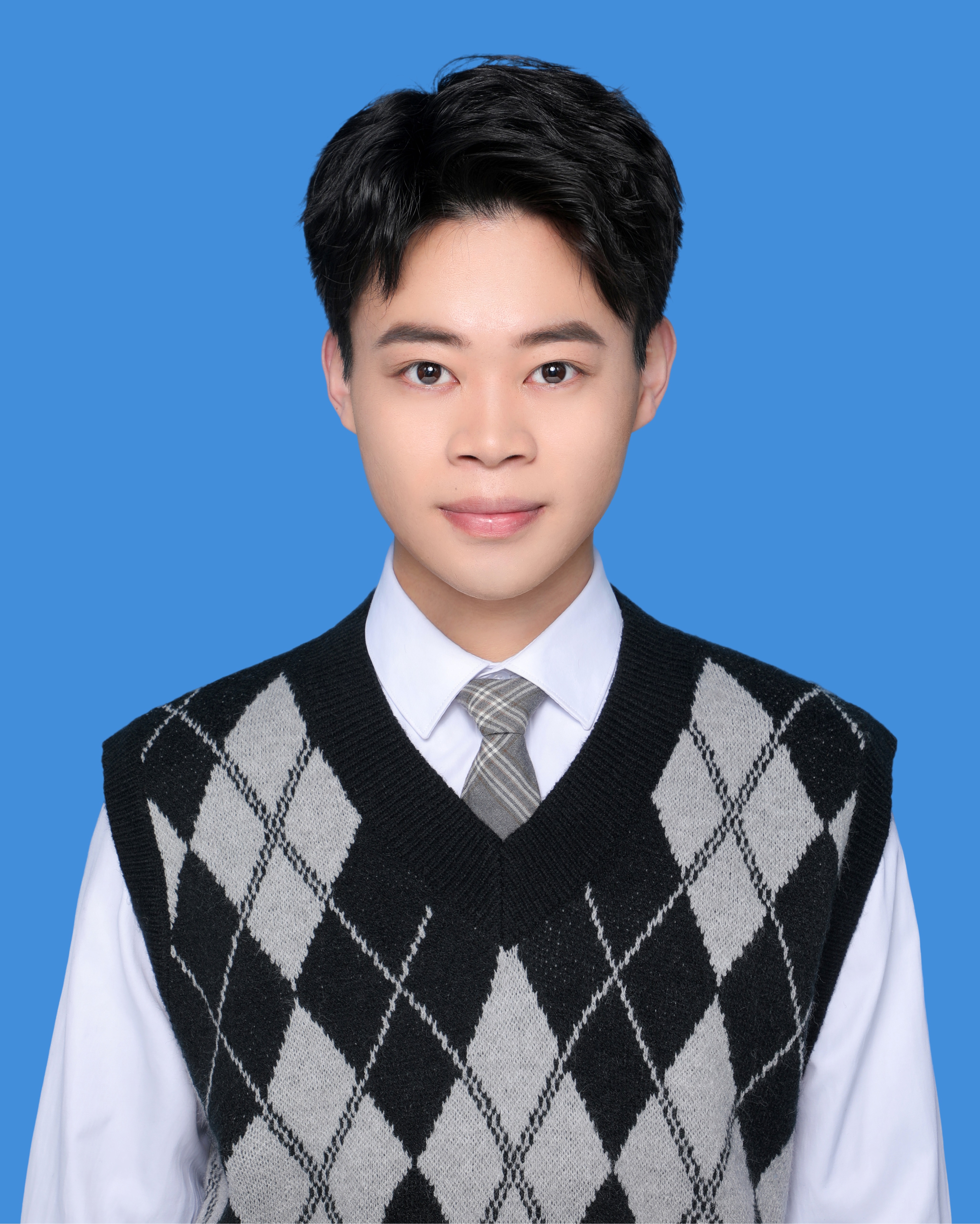}}]{Xin Liu}
	received his Bachelor degree in Automation from Yanshan University in 2019. He is currently pursuing toward the Ph.D. in the Department of Electrical Engineering of Yanshan University. His research interest covers SLAM, Edge-Computing and multi-sensor information fusion.
\end{IEEEbiography}
\vspace*{-1.1cm}

\begin{IEEEbiography}[{\includegraphics[width=1in,height=1.25in,clip,keepaspectratio]{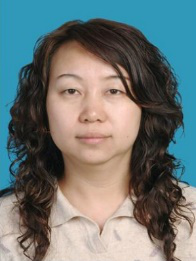}}]{Shuhuan Wen}
	was received the PhD degree in control theory and control engineering from the Yanshan University, Qinhuangdao, China in 2005. She is currently a Professor of automatic control in the Department of Electric Engineering, Yanshan University. She has coauthored one book, about 40 papers. Her research interests include SLAM, computer vision, robotics, 3D object recognization and reconstruction. 
	Dr. Wen was a Visiting Scholar of the Ottawa University, Carleton University and Simon Fraser University in Canada in 2011-2013. She is also a visiting professor at University of Alberta in Canada in 2021-2022. She is an AE of IROS in 2021 to 2022.
\end{IEEEbiography}
\vspace*{-1.1cm}
\begin{IEEEbiography}[{\includegraphics[width=1in,height=1.25in,clip,keepaspectratio]{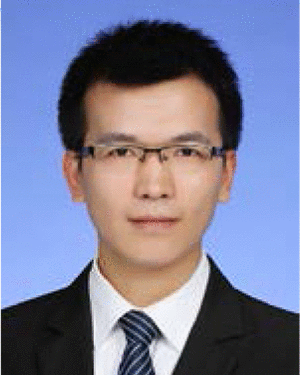}}]{Jing Zhao}
	is an associate professor at the Department of Automation, Yanshan University, Qinhuangdao, P. R. China. He received a B.S. in Automation from Shandong University, Jinan, P.R. China in 2008, a M.S. in measuring and testing technologies and instruments from 4th Academy of China Aerospace Science and Technology Corporation (CASC), Xian, P.R. China in 2011, and a Ph.D. in control science and engineering from Tianjin University, Tianjin, P.R. China in 2016. His research interests include brain-computer interface and brain robot cooperation.
\end{IEEEbiography}
\vspace*{-1.1cm}
\begin{IEEEbiography}[{\includegraphics[width=1in,height=1.25in,clip,keepaspectratio]{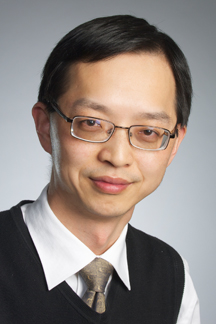}}]{Tony Z. Qiu}
	received the B.S. degree in control science and engineering and the M.S. degree in intelligent transportation system both from Tsinghua University, Beijing, China, in 2001 and 2003, respectively, and the Ph.D. degree from the University of Wisconsin, Madison, in 2007. From 2008 to 2009, he worked as a Postdoctoral Researcher with California Partners for Advanced Transit and Highways (PATH), University of California, Berkeley. He is currently working as an Assistant Professor with the Department of Civil and Environmental Engineering, University of Alberta, Edmonton, AB, Canada. His research interest include intelligent transportation systems, traffic modeling and simulation, traffic operation and control, traffic state estimation and prediction, and safety analysis for urban and freeway.
\end{IEEEbiography}
\vspace*{-1.2cm}
\begin{IEEEbiography}[{\includegraphics[width=1in,height=1.25in,clip,keepaspectratio]{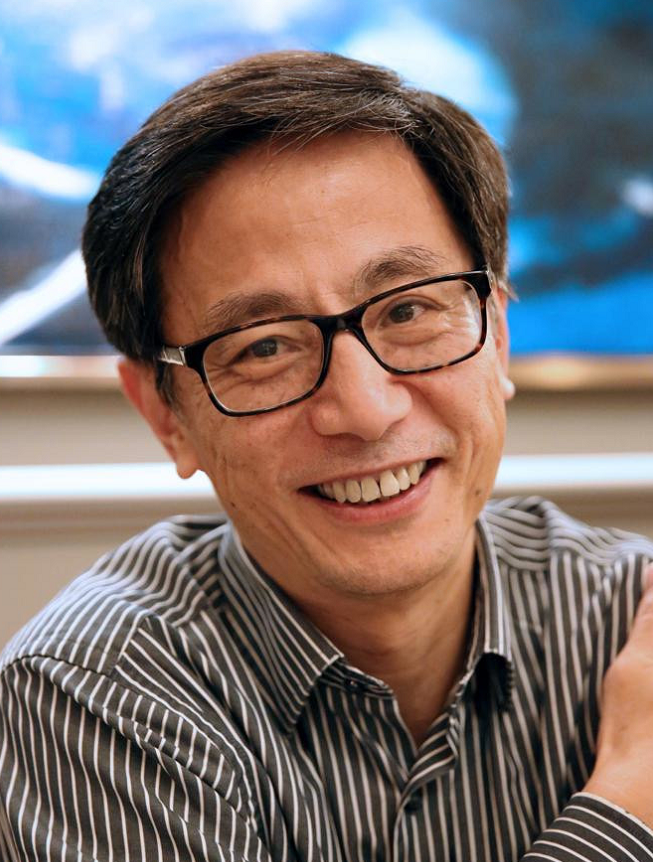}}]{Hong Zhang}
	was received the B.Sc degree from Northeastern University, Boston, USA, in 1982, and the PH.D.degree from Purdue University. West Lafayette, IN, USA, in 1986, both in electrical engineering. He conducted post-doctoral research at the University of Pennsylvania from 1986 to 1987 before joining the Department of Computing Science, University of Alberta, Canada. He is an NSERC Industrial Research Chair. His current research interests include robotics computer vision, and image processing.
\end{IEEEbiography}
\vfill

\end{document}